\documentclass[runningheads]{llncs}
\usepackage{graphicx}
\usepackage{graphics}
\usepackage{comment}
\usepackage{amsmath,amssymb} 
\DeclareMathOperator*{\argmax}{argmax}
\usepackage{color}
\usepackage{caption}
\usepackage{subcaption}
\usepackage{bbm}
\usepackage{multirow}
\usepackage[symbol]{footmisc}
\usepackage{float}

\newcommand{\etal}{\textit{et al.}}

\begin{document}
\pagestyle{headings}
\mainmatter
\def\ECCVSubNumber{6436}  
\newcommand{\steph}[1]{\textcolor{red}{Steph: #1}}
\newcommand{\eli}[1]{{\color{magenta}#1}}
\newcommand{\willi}[1]{\textcolor{blue}{#1}}
\newcommand{\methodname}{ACIDS}

\title{Learning to Cluster under Domain Shift} 



\titlerunning{Learning to Cluster under Domain Shift}
%
\author{Willi Menapace\inst{1} \and 
St\'{e}phane Lathuili\`{e}re\inst{3} 
\and
Elisa Ricci\inst{1,2}}
\authorrunning{W. Menapace et al.}
%
\institute{University of Trento, Trento, Italy \and
Fondazione Bruno Kessler, Trento, Italy
\and
LTCI, T\'{e}l\'{e}com Paris, Institut Polytechnique de Paris, Palaiseau, France\\
\email{willi.menapace@gmail.com}\\
}

\maketitle

\begin{abstract}
While unsupervised domain adaptation methods based on deep architectures have achieved remarkable success in many computer vision tasks, they rely on a strong assumption, i.e. labeled source data must be available. In this work we overcome this assumption and we address the problem of transferring knowledge from a source to a target domain when both source and target data have no annotations. Inspired by recent works on deep clustering, our approach leverages information from data gathered from multiple source domains to build a domain-agnostic clustering model which is then refined at inference time when target data become available. Specifically, at training time we propose to optimize a novel information-theoretic loss which, coupled with domain-alignment layers, ensures that our model learns to correctly discover semantic labels while discarding domain-specific features. Importantly, our architecture design ensures that at inference time the resulting source model can be effectively adapted to the target domain without having access to source data, thanks to feature alignment and self-supervision. We evaluate the proposed approach in a variety of settings\footnote[1]{Code available at github.com/willi-menapace/acids-clustering-domain-shift}, considering several domain
adaptation benchmarks and we show that our method is able to automatically discover relevant semantic information even in presence of few target samples and yields state-of-the-art results on  multiple domain adaptation benchmarks. 
\end{abstract}

\section{Introduction}
\begin{figure}[t]
\centering
\includegraphics[width=0.99\textwidth]{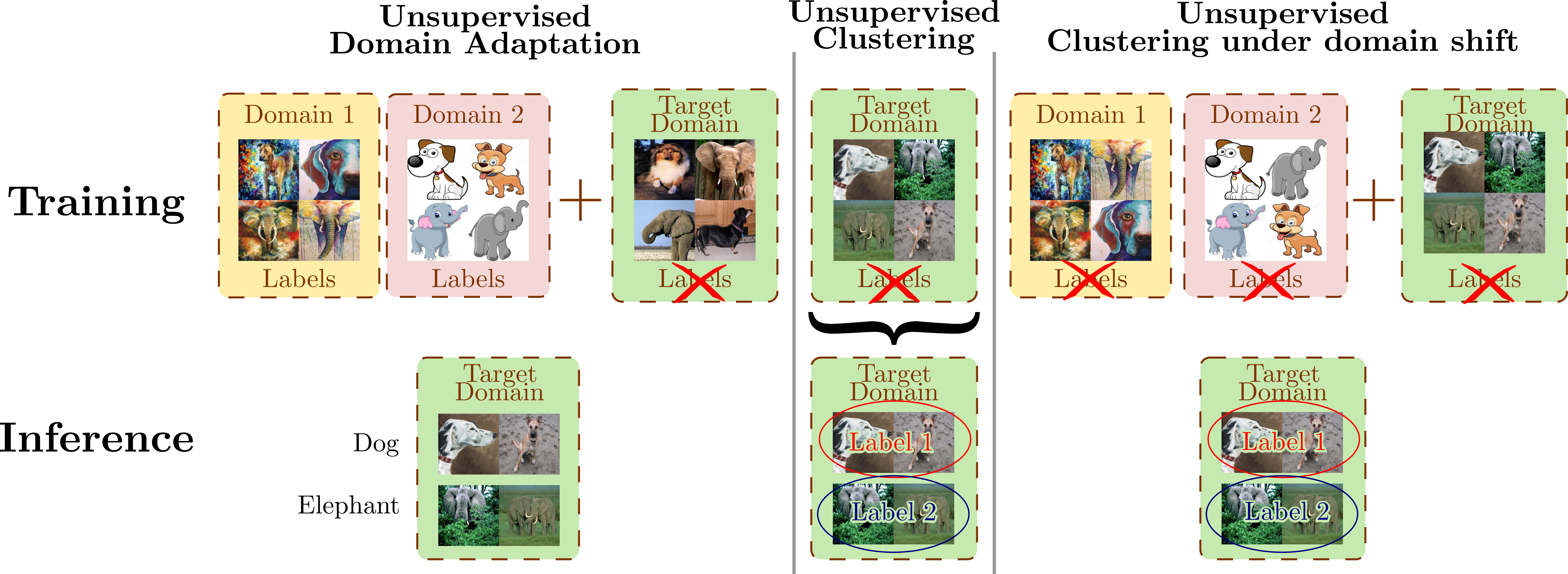}
\caption{In the Unsupervised Domain adaptation (UDA) setting, a model is trained combining labeled images from one or several source domains and unlabeled images from the target domain. In the unsupervised clustering setting, unlabeled images from the same domain are grouped into visually similar images. We introduce the Unsupervised Clustering under Domain Shift (UCDS) setting where we leverage unlabeled source domain data to improve target domain clustering.}

\label{fig:teaser}
\end{figure}

The astonishing performance of deep learning models in a large variety of applications must be partially ascribed to the availability of large-scale datasets with abundant annotations. 
Over the years, several solutions have been proposed to avoid prohibitively
expensive and time-consuming data labeling such as transfer learning \cite{pan2011domain} or domain adaptation \cite{csurka2017domain} strategies. In particular, unsupervised domain adaptation (UDA) methods \cite{long2015learning,long2017deep,morerio2017minimal,peng2018synthetic,carlucci2017autodial,li2016revisiting,roy2019unsupervised,hoffman2017cycada,zen2014unsupervised}
leverage the knowledge extracted from labeled data of one (or multiple)
source domain(s) to learn a prediction model for a different but related target domain where no labeled data are available. This strategy is illustrated in Fig.\ref{fig:teaser}-left.

Over the last decade, increasing efforts have been devoted
to develop deep architectures for UDA and promising results have been obtained in several applications such as object recognition \cite{long2015learning,tzeng2015simultaneous,carlucci2017autodial}, semantic
segmentation \cite{hoffman2017cycada}, depth estimation \cite{zhao2019geometry}, etc.
While effective in many tasks, current UDA methods rely on a key assumption: annotations associated with data from the source domain(s) must be available. In this paper, we argue that this assumption may hinder the use of UDA in many practical applications. For instance, relaxing the constraints of disposing of labeled source data can broaden the applicability of knowledge transfer methods to tasks and scenarios where gathering annotations is challenging or even impossible (e.g. medical). 

A possible alternative to supervised training is unsupervised clustering (Fig.\ref{fig:teaser}-center). Clustering is a class of unsupervised learning methods that are designed to group images in such a way that images in the same group contain similar content. 
Recently, some works \cite{DBLP:journals/corr/abs-1807-05520,haeusser,DBLP:journals/corr/abs-1807-06653} have shown that appropriately designed deep architectures can be successfully used to discover clusters in a training set and perform representation learning. By opposition to UDA, clustering does not require any annotation. However, it relies on the assumption that all the data belongs to the same domain. If this condition is not fulfilled, clustering algorithms would tend to group data according to the visual style associated to their domain and not according to their semantic content.

Motivated by these observations, in this paper, we propose a new setting, Unsupervised Clustering under Domain Shift (UCDS), (see Fig.\ref{fig:teaser}-right) where we assume that we dispose of data from different known domains but no class labels are available in both source and target domains.
Our approach develops under the assumption that, while no annotations from source data are available, still we may benefit from the access to multiple datasets, i.e. to multiple source domains. This is very reasonable as in many practical applications it is very likely to dispose of several datasets collected under different conditions.

Our method develops from the intuition that, by combining multiple domains with different visual styles, we can obtain clusters based on the semantic content rather than on stylistic or texture features. Importantly, by leveraging multiple source domains, we show that the target domain can be clustered accurately even when target data is limited.
Our method is organized in two steps. First, a novel multi-domain deep clustering model is learned which, by seamlessly combining domain-specific distribution alignment layers \cite{carlucci2017autodial} and an information-theoretic loss permits to discover semantic categories across domains. In a subsequent step, target data are exploited to refine the learned clustering model
by simultaneously matching source features distributions with domain-alignment layers and by maximizing the mutual information between the class assignments of pairs of perturbed samples.
Recalling these elements, we name our algorithm \methodname: Adaptive Clustering of Images under Domain Shift.

The major advantage of our two-stage pipeline is that it does not require source and target data to be available simultaneously. Consequently, our setting differs from classical UDA and unsupervised transfer learning scenarios~\cite{csurka2017domain,pan2011domain} since only the source model is provided to the unlabeled target domain. Discarding the source data at adaptation time can broaden the applicability of our framework to tasks and scenarios that suffer from transmission or privacy issues. 
Our extensive experimental evaluation demonstrates that our approach successfully discovers semantic categories and outperforms state of the art unsupervised learning models on popular domain adaptation benchmarks: Office-31~\cite{saenko2010adapting}, PACS~\cite{li2017deeper} and Office-Home \cite{venkateswara2017deep} dataset.

\paragraph{Contributions.} To summarize, the main contributions of this work are: (i) We introduce a new setting, Unsupervised Clustering under Domain Shift (Fig.\ref{fig:teaser}-right), where we learn a semantic predictor from unsupervised target samples leveraging from multiple unlabeled source domains; (ii) We propose an information-theoretic algorithm for unsupervised clustering that operates under domain shift. Our method successfully integrates the data-augmentation strategy typically used by deep clustering methods \cite{DBLP:journals/corr/abs-1807-06653} within a feature alignment process; (iii) We evaluate our method on several domain adaptation benchmarks demonstrating that our approach can successfully discover semantic categories even in the presence of domain shift and with few target samples.

\section{Related works}

In the following we review previous approaches on UDA, discussing both single source and multi-source  methods. Since we propose a deep architecture for unsupervised learning under domain shift, we also review related work on deep clustering.

\noindent\textbf{Domain adaptation}. Earlier UDA methods assume that only a single source domain is available for transferring knowledge. These methods can be roughly categorized into three main groups. The first category includes methods which align source and target data distributions  by matching the distribution statistical moments of different orders. For instance, Maximum Mean Discrepancy, \textit{i.e.} the distance between the mean of domain feature distributions, is considered in \cite{long2015learning,long2017deep,venkateswara2017deep,tzeng2014deep}, while second order statistics are used \cite{sun2016deep,morerio2017minimal,peng2018synthetic}. Domain alignment layers derived from batch normalization (BN) \cite{ioffe2015batch} or whitening transforms \cite{siarohin2018whitening} are employed in \cite{carlucci2017autodial,li2016revisiting,mancini2018boosting,roy2019unsupervised}.

The methods in the second category learn domain-invariant representations considering an adversarial framework. For instance, in \cite{ganin2014unsupervised} a gradient reversal layer is used to learn domain-agnostic representations.  Similarly, ADDA  \cite{ADDA} introduces a domain confusion loss to align the source and the target domain feature distributions. The third category of methods consider a generative framework (i.e., GANs (\cite{Goodfellow:GAN:NIPS2014}) to create synthetic source and/or target images. Notable works are 
CyCADA \cite{hoffman2017cycada}, I2I Adapt \cite{murez2018image} and Generate To Adapt (GTA) \cite{sankaranarayanan2018generate}. Our method is related to previous works in the first category, as we also leverage domain-alignment layers to perform adaptation. However, we consider a radically different setting where no annotation is provided in the source domain and only the source model (and not the source data) is exploited at adaptation time. 


While most previous works on UDA consider a single source domain, recently some works have shown that performance can be considerably improved by leveraging multiple datasets. For instance, in \cite{mancini2018boosting} multiple latent source domains are discovered and used for transferring knowledge. Recently, Deep Cocktail Network (DCTN) \cite{xu2018deep} introduce a distribution-weighted rule for classification which is combined with an adversarial loss. $\textrm{M}^{3}\textrm{SDA}$ is described in \cite{peng2018moment}: it reduces the discrepancy between the multiple source and the target domains by dynamically aligning moments of their feature distributions. 

Differently from these methods, \methodname~does not assume annotations in the source domain. One related work to ours is \cite{mancini2019adagraph} where information from multiple source domains is exploited for constructing a domain-dependency graph and then used when the target data are made available. However, in \cite{mancini2019adagraph} an entropy loss for target model adaptation is considered, which we experimentally observe is less effective than our proposal self-supervised loss.
Our method is also related to recent domain generalization (DG) methods \cite{carlucci2019domain,li2019episodic}. In fact, similarly to DG, we also assume that source and target data are not simultaneously available. However, differently from DG, we make use of target data for model adaptation when they are available.

\noindent\textbf{Deep Clustering.} Over the last few years, unsupervised representation learning has attracted considerable attention in the computer vision community. Self-supervised learning approaches mostly differ in the self-supervised losses used to learn feature representations. Notable examples are methods which derive indirect auxiliary supervision from spatial-temporal  consistency \cite{wang2017transitive}, from solving  jigsaw puzzles \cite{noroozi2016unsupervised} or from colorization \cite{larsson2016learning}. 

Recently, some studies have attempted to derive deep clustering algorithms which simultaneously discover groups in training data and perform representation learning. For instance, DEC \cite{xie2015unsupervised} makes use of an autoencoder to produce a latent space where cluster centroids are learned. DAC \cite{chang2017dac} casts the clustering problem into pairwise-classification using a convolutional network to learn feature representations. In \cite{DBLP:journals/corr/abs-1807-05520} {DeepCluster}, an iterative clustering procedure is devised which adopts k-means to learn representations and uses the subsequent assignments as supervision. 
Similarly, in \cite{haeusser} an end-to-end clustering approach is proposed where an encoder network is trained with an alternate scheme.
Recently, Ji \etal \ propose Invariant Information Clustering (IIC) \cite{DBLP:journals/corr/abs-1807-06653}, where a deep network is learned with an information-theoretic criterion in order to output semantic labels, rather than high dimensional representations. Our approach is inspired by this method. However, we specifically address the problem of transferring knowledge from source to target domains.

\section{Proposed Method}

\begin{figure}[t]\centering
\begin{subfigure}[b]{0.5\textwidth}
         \centering
\includegraphics[height=5.2cm]{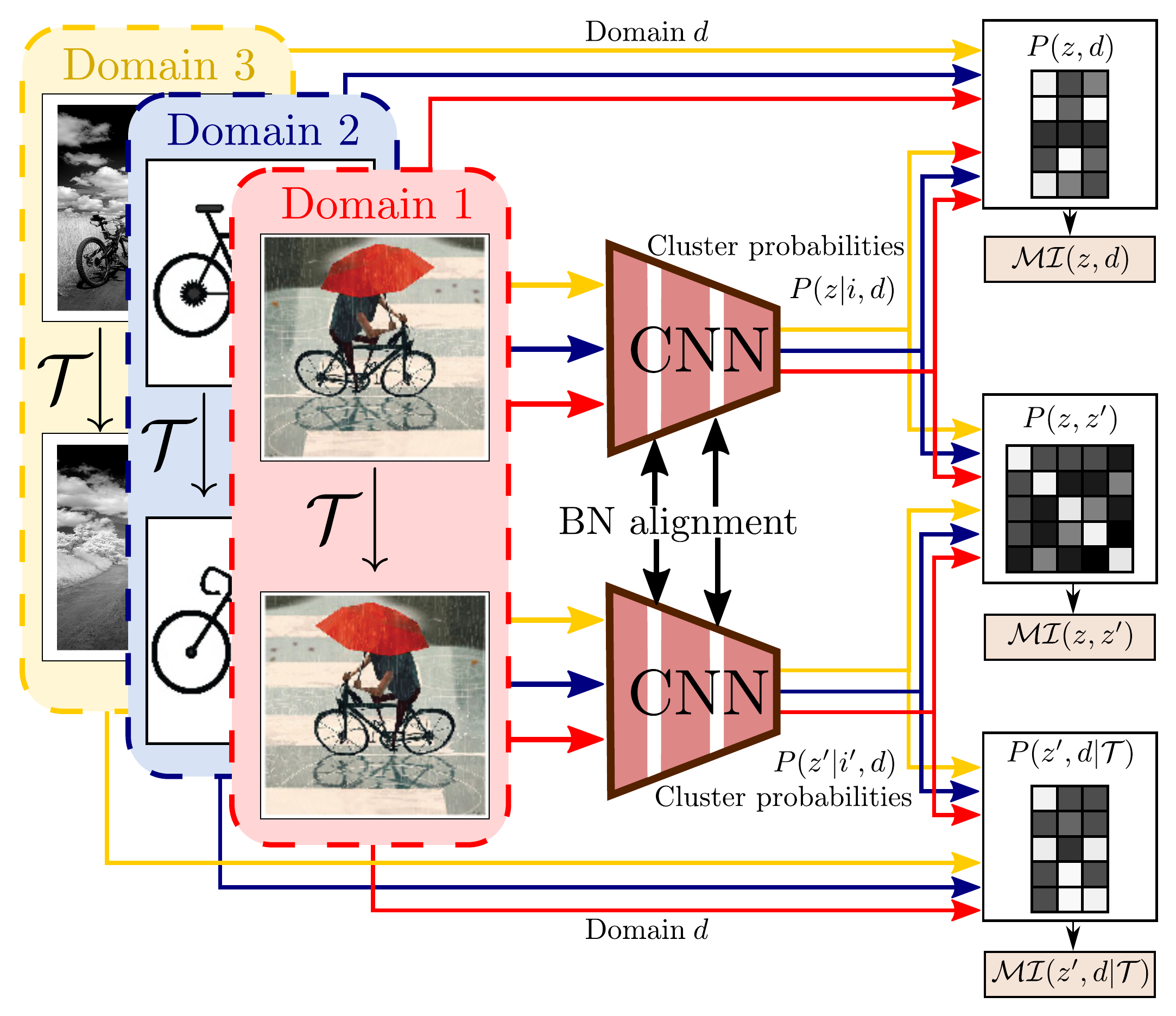} 

\caption{Training on Sources\label{fig:train}}
     \end{subfigure}\hfill
\begin{subfigure}[b]{0.4\textwidth}
         \centering
\includegraphics[height=5.2cm]{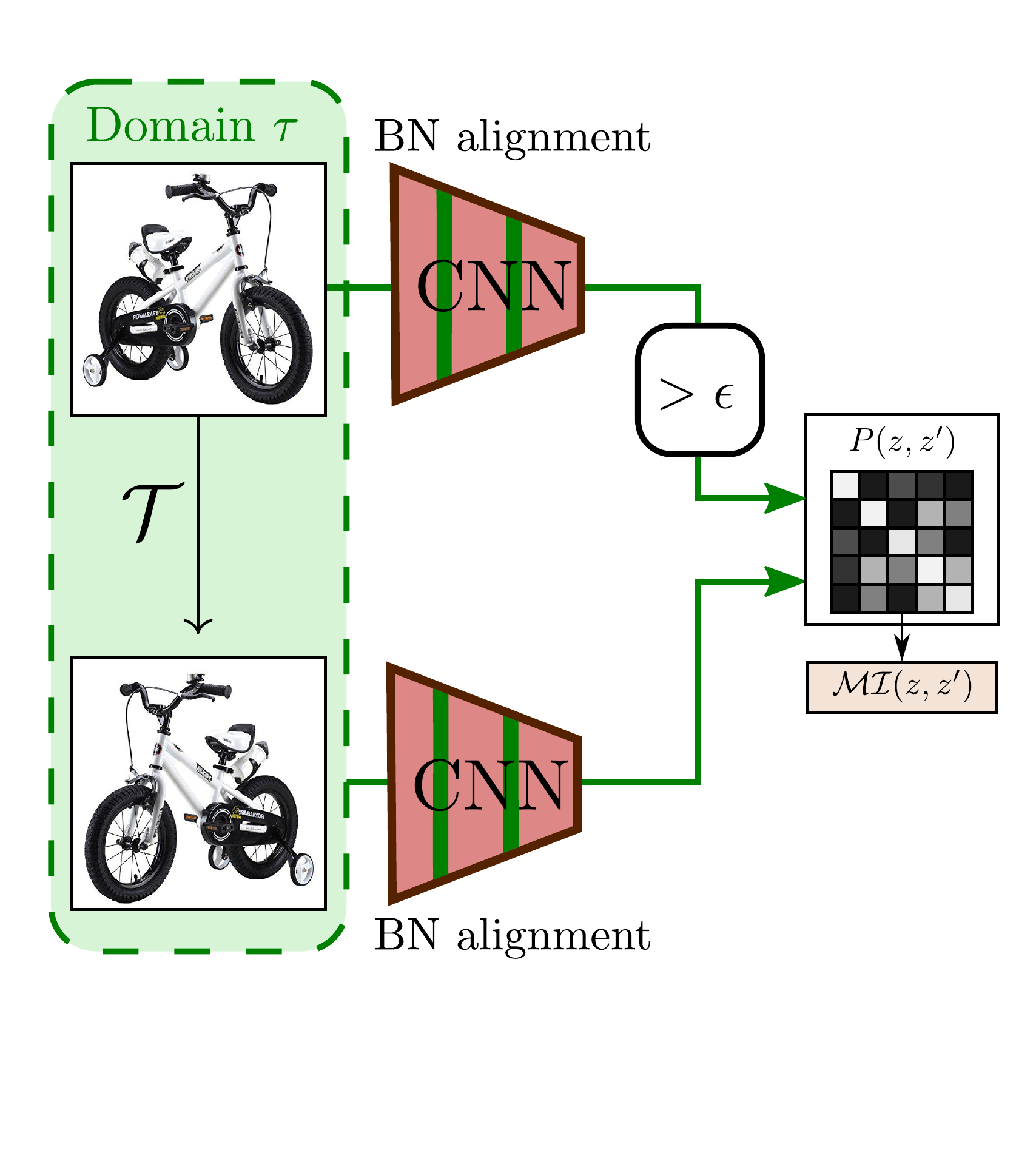} 

\caption{Adaptation on Target\label{fig:ada}}
\end{subfigure}

\caption{Illustration of the \methodname~framework for UCDS. In the training stage (Fig.\ref{fig:train}), images are clustered by maximizing mutual information between the predictions from the original and transformed images. Domain alignment is addressed combining a Batch Normalization (BN) alignment technique with a novel mutual information minimizing formulation. In the adaptation phase (Fig.\ref{fig:ada}), domain shift is handled by combining BN alignment with a specific mutual information maximization procedure.}

\label{fig:pipeline}
\end{figure}


In this section, we introduce the proposed \methodname~fully unsupervised multi-source domain adaptation framework. The design of the source training framework is guided by two motivations. First, we need to take advantage of the different domains in order to obtain clusters that correspond to semantic labels rather than domain-specific image styles. Second, the network must learn image representations that can be transferred to any unknown target domain.

We assume to observe $S$ source domains $I_s$ composed of images, each depicting an object from $C$ different object categories. In this work, we assume that $C$ is known a priori but we do not dispose of image labels.
We propose to learn image representations allowing to cluster source images according to the unknown category labels. Our goal is to adapt the representation learned on the source domains in order to predict labels on a target domain $I_t$. In this adaptation stage, we consider that we do not dispose of the images from $I_1, \dots, I_S$. To this aim, we employ a deep neural network $\phi_\theta :I \rightarrow Z$ with parameters $\theta$ that predicts cluster assignments probabilities.
To obtain network outputs $Z\in[0,1]^C$ that can be interpreted as probability vectors, $\phi_\theta$ is terminated by a layer with a softmax activation function.  

In Sec.\ref{sec:multiIIC} we describe the objective used to cluster source images. In order to ensure clustering based on semantic labels and not on domain-specific styles, we introduce in Sec.\ref{sec:domainAlignment} a novel information-theoretic alignment mechanism based on the minimization of mutual information between domains and cluster assignments. Here, we also detail our batch normalization alignment layers that complement the framework. Adaptation on the target domain is described in Sec.\ref{sec:ada}.

\subsection{Multi-domain Clustering with Mutual information}
\label{sec:multiIIC}

Let $i\in I=\bigcup\limits_{s=1}^{S} I_s$ be an image from the domain $d\in{1..S}$. Both $i$ and $d$ are treated here as random variables. We consider that we dispose of a set of image transformations $\mathcal{T}$. After sampling a transformation $t\in \mathcal{T}$,
we obtain a transformed version of the image $i$ denoted as $i'$. Following the approach of \cite{DBLP:journals/corr/abs-1807-06653}, we train $\phi_\theta$ in such a way that, first, it returns the same output for both $i$ and $i'$ and, second, it returns different outputs for different images. This double objective can be achieved by maximizing the mutual information between the predictions from $i$ and $i'$ with respect to the network parameters:

\begin{align}
  \max_{\theta}\mathcal{MI}(z,z')
\end{align}

where $z=\phi_\theta(x)$ and $z'=\phi_\theta (x')$ are the network cluster assignment predictions.
To estimate the mutual information $\mathcal{MI}(z,z')$, we need to compute the joint probability of the cluster assignment $P_{cc'}=P(z\!=\!c, z'\!=\!c')$ where $c\in{1..C}$ and $c'\in{1..C}$ are all the possible cluster indexes. This probability is estimated by marginalization over the current batch. Let us assume to observe a batch composed of $N$ unlabeled images $\{i_s^n\}_{n=1}^{N}\subset I_s$ from each of $S$ source domains. We have:

\begin{align}
  P_{cc'}&=P(z\!=\!c, z'\!=\!c')=\sum_{s=1}^S P(d\!=\!s)P(z\!=\!c, z'\!=\!c'|d\!=\!s)\notag\\
  &=\frac{1}{SN}\sum_{s=1}^S\sum_{n=1}^N \phi_\theta(i_s^n).\phi_\theta({i_s^n}')^\top\label{eq:Pcc}
\end{align}
Similarly, we estimate the marginal distribution: 
\begin{align}
P_c=P(z\!=\!c)=\frac{1}{S}\sum_{s=1,n=1}^{S\times N} \phi_\theta(i_s^n) ~\text{and}~ P_c'=P(z\!=\!c')=\frac{1}{S}\sum_{s=1,n=1}^{S\times N} \phi_\theta({i_s^n}')\notag.
\end{align}
From these probability distributions, the mutual information loss is given by:
\begin{align}
  \mathcal{MI}(z,z')=\sum_{c=1}^C\sum_{c'=1}^C P_{cc'} \ln\frac{P_{cc'}}{P_{c}.P_{c'}}
  \label{eq:MIzz}
\end{align}

\subsection{Domain Alignment}
\label{sec:domainAlignment}

\noindent\textbf{Feature alignment via mutual information minimization.}
Training the network $\phi_\theta$ only via the maximization of \eqref{eq:MIzz} may lead to solutions where input images are clustered according to the domain information rather than their semantics. To tackle this problem, feature distribution from the different domains should be aligned in such a way that the classifier cannot cluster images according to the domain. We propose to address this domain alignment problem by the combination of two complementary strategies.

First, we propose to formulate domain alignment as a mutual information optimization problem. The key idea of \methodname~alignment strategies is that cluster assignment $z$ should be independent from the domain $d$ of the input image. Consequently, the mutual information between the predicted label $z$ and the image domain $d$ must be minimal. To this aim, we estimate the joint probability distribution $P(z, d)$ by marginalization:

\begin{align}
\label{eq:miDomainAlignment}
  P(z, d)=\frac{1}{N}\sum_{s=1}^S\sum_{n=1}^N \mathbbm{1}(s=d)\phi_\theta(i_s^n)
\end{align}
Similarly to \eqref{eq:MIzz}, we can estimate $\mathcal{MI}(z,d)$. Note that this mutual information loss leads to an extremely limited computation overhead compared to alternative solutions such as adversarial approaches. 

Even though minimizing $\mathcal{MI}(z,d)$ enforces alignment between the domains, this formulation does not take advantage of the image transformation framework described in Sec.\ref{sec:multiIIC}. In order to further use the potential of our data augmentation approach, we propose to use the transformed image to favor domain alignment. More specifically, for every transformation $t\in\mathcal{T}$, the cluster assignment $z'$ should be independent from the domain $d$ of the input image. In other words, the mutual information $\mathcal{MI}(z',d|t)$ should be minimized for every transformation $t$. Here again, the mutual information $\mathcal{MI}(z',d|t)$ is computed via marginalization similarly to $\mathcal{MI}(z,z')$ and  $\mathcal{MI}(z,d)$.

This mutual information minimization is both lightweight and efficient. Nevertheless, it acts only according to a top-down strategy, since alignment is imposed only on the output of the network and not in the early layers. Consequently, we propose to complement our framework with a feature alignment strategy based on batch normalization that acts all over the network.

\noindent\textbf{Feature alignment via batch normalization.}
We consider that the network $\phi_\theta$ embeds Batch normalization (BN) layers. We adopt the idea of previous works \cite{li2016revisiting,mancini2018kitting,carlucci2017autodial} and perform domain adaptation by updating the BN statistics. The main assumption behind this strategy is that the domain-shift can be reduced by aligning the different source feature distributions to a Gaussian reference distribution. We consider that we observe $S$ batches of images, one from each of the $S$ source domains. Assuming a given BN layer, $\{x_s^n\}_{n=1}^{N}$ and $\{{x_s^n}'\}_{n=1}^{N}$ denote the features, corresponding to domain $s$, in input to the BN layer for each image and the transformed counterpart respectively. We compute the batch statistics for each domain separately:

\begin{align}
\forall s \in \{1..S\},\hspace{0.3cm} \hat{\mu}_s=\frac{1}{2m}\sum_{n=1}^{N}(x_s^n + {x_s^n}')  \hspace{0.3cm}  \hat{\sigma}_s^2=\frac{1}{2m}\sum_{n=1}^{N}[(x_s^n-\hat{\mu}_s)^2 + ({x_s^n}'-\hat{\mu}_s)^2] \label{eq:musigma}
\end{align}

For a given input $x$ computed from an image of the domain $s$, the output of the normalization layer is computed as follows:
\begin{align}
\hat{x}=\gamma \frac{x-\mu_s}{\sqrt{\sigma_s^2+\epsilon}}+\beta\label{eq:BNoutput}
\end{align}
where $\gamma$ and $\beta$ are the usual affine transformation parameters of the BN layer, while $\epsilon\in\mathbb{R}$ is a
constant introduced for numerical stability. Note that the affine transformation parameters are shared among the different domains.
This strategy guarantees that every BN layer outputs feature distributions from every domain with a mean value equal to 0 and a variance equal to 1. The main advantage of \methodname~framework is twofold. First, it does not require any additional loss that would imply more hyper-parameter tuning to obtain good convergence. Second, adaptation on the target data can be performed without accessing the source data. 

\subsection{Training and adaptation procedures}
\label{sec:targetAdaptation}
\noindent\textbf{Overall objective function.}
\noindent
In the previous section, we detailed how we estimate three different mutual information terms. The term $\mathcal{MI}(z,z')$ must be maximized while $\mathcal{MI}(z,d)$ and $\mathcal{MI}(z',d|t)$ must be minimized. Consequently the total minimization objective function can be written:
\begin{align}
  \mathcal{L}=-\mathcal{MI}(z,z')+\mathcal{MI}(z,d)+\frac{1}{T}\sum_{t\in\mathcal{T}}\mathcal{MI}(z',d|t).
\end{align}
where $T$ is that cardinality of $\mathcal{T}$.

\noindent\textbf{Improving stability.}
\noindent
The computation of the mutual information $\mathcal{MI}(z,z')$ is based on the estimation of the marginal probability matrix $P_{cc'}\in\mathbb{R}^{C\times C}$. Following a standard SGD approach, this matrix is computed for every batch. However, estimating this full probability matrix from a small batch can be inaccurate when the number of classes $C$ is high. In addition, increasing the batch size may lead to memory issues. Practically, we observed in our preliminary experiments, that a large batch size is critical to obtain satisfying convergence of the IIC model. In our context, the issue appears to be even stronger since the images originate from different domains. Our assumption is that the higher variance of the features, despite feature alignment, leads to gradients with higher variance and unstable training. To tackle this issue, we propose to robustify the estimation of the marginal probability matrices using a moving average strategy. Considering that $P_{cc'}$ is the matrix associated to the current batch, the mutual information in Eq.\eqref{eq:MIzz} is computed using $\tilde{P}_{cc'}$:
\begin{align}
\tilde{P}_{cc'}=\alpha P_{cc'}+ (1-\alpha)\hat{P}_{cc'}
\end{align}
where $\hat{P}_{cc'}$ is the probability matrix $\tilde{P}_{cc'}$ estimated on the previous batch and $\alpha$ is a dynamic parameter. From a probabilistic point of view, this formulation can be understood as a stronger marginalization since the distribution is estimated considering in Eq.\eqref{eq:Pcc} not only on the N samples of the current batches but also the past batches. This estimation is correct under the assumption that the network $\phi_\theta$ did not change too much in past SGD steps.

\noindent\textbf{Adaptation to the target domain.}
\label{sec:ada}
\noindent
At test time, we dispose of images from the target domain $\{i_\tau^n\}_{n=1}^{N_\tau}\in I_\tau$. However, we assume that we do not dispose anymore of the training data from the source domains. Adaptation is performed using two successive procedures. First, in order to align the feature distribution of the target data with the source distributions, we estimate the statistics of the inputs of each BN layer as in Eq.\eqref{eq:musigma}. The output of each BN layer is then computed according to Eq.\eqref{eq:BNoutput}. Second, our model is adapted using a variant of the mutual formulation used at training time and described in Sec.\ref{sec:multiIIC} computed only on the target domain $I_\tau$. We argue that in an unsupervised setting it is beneficial to treat samples with high prediction confidence differently from the ones with low confidence \cite{sohn2020fixmatch}. The rationale is to drive low confidence predictions towards certainty represented by high confidence predictions while not altering the latter.
We propose to treat images $i_\tau$ with a prediction confidence larger than a given threshold $\epsilon$ as fixed points whose output class prediction $c$ must be replicated by the corresponding transformed image $i'_\tau$. This differs from the mutual information approach employed in Sec.\ref{sec:multiIIC} where the output correspondence is achieved only implicitly and an incorrect class assignment to image $i'_\tau$ may negatively alter the prediction of $i_\tau$ as well, causing instability.
We define:

\begin{equation*}
    \tilde{\phi}_\theta(i) =
    \begin{cases}
      \mathbbm{1}(c = \argmax \phi_\theta(i)) & \text{if} \max \phi_\theta(i) \geq \epsilon\\
      \phi_\theta(i) & \text{otherwise}\\
    \end{cases}  
\end{equation*}
\begin{align*}
  P_{cc'}=\frac{1}{N}\sum_{n=1}^N \tilde{\phi}_\theta(i^n_\tau).\phi_\theta({i^n_\tau}')^\top\label{eq:PccFixmatch} ~\text{and}~ P_c=P(z\!=\!c)=\sum_{n=1}^{N} \tilde{\phi}_\theta(i^n_\tau) 
\end{align*}

Then, we compute the mutual information term $\mathcal{MI}(z,z')$ in Eq.\ref{eq:MIzz} using the newly defined $P_{cc'}$ and $P_c$. 
This newly defined mutual information loss no longer suffers from the wrong $i'_\tau$ prediction problem because the $\argmax$ operation stops gradient propagation in the high confidence predictions, fixing them and making the model focus on low confidence ones. 


\section{Experiments}

\begin{table}[t]
\begin{center}
\caption{Ablation results on the PACS dataset: (i) training is performed on a single, merged source domain (ii) training performed on a single source domain, (iii) removed feature alignment via mutual information minimization, (iv) removed BN feature alignment + (iii); (v) no target adaptation, (vi) target adaptation using entropy instead of mutual information. Accuracy (\%) on target domain.}
\label{table:pacsAblation}
\footnotesize
\begin{tabular}{lccccc}
\hline\noalign{\smallskip}
Target domain: & A & C & P & S & Avg\\
\hline\noalign{\smallskip}
Merged source (i) & 23.6 & 32.8 & 31.2 & 28.2 & 29.0 \\
\hline
\noalign{\smallskip}

Single source A (ii) & - & 31.0 & 45.8 & 28.7 & - \\
Single source C (ii) & 31.0 & - & 42.5 & 35.0 & - \\
Single source P (ii) & 33.0 & 33.8 & - & 30.5 & - \\
Single source S (ii) & 25.0 & 30.2 & 37.2 & - & - \\
\hline\noalign{\smallskip}
w/o domain mi loss (iii) & 27.4 & 24.3 & 50.9 & 23.0 & 31.4 \\
w/o BN alignment (iv) & 23.7 & 34.3 & 38.6 & 23.0 & 22.1 \\
\hline\noalign{\smallskip}
w/o target adaptation (v) & 34.8 & 36.5 & 44.2 & 40.8 & 39.1 \\
entropy target adaptation (vi) & 29.2 & 36.6 & 29.7 & 41.0 & 34.1 \\
\hline\noalign{\smallskip}
\methodname & \textbf{42.1} & \textbf{44.5} & \textbf{64.4} & \textbf{51.1} & \textbf{50.5} \\

\hline
\end{tabular}
\end{center}

\end{table}
In this section, we evaluate the effectiveness of \methodname~on three widely used domain adaptation datasets and perform an ablation study showing the importance of each component of our method.

\noindent \textbf{Datasets.}
The PACS \cite{li2017deeper} dataset is a domain adaptation dataset composed of 9,991 images divided in 7 classes spanning 4 different domains: Photo (P), Art (A), Cartoon (C) and Sketch (S). The different domains of PACS represent a rich variety of visual characteristics, from natural images to sketches, which cause large semantic gaps and make it a challenging domain adaptation dataset.

The Office31 dataset \cite{saenko2010adapting} contains 4,110 images divided in 3 different domains and 31 classes, namely: Amazon (A), DSLR (D) and Webcam (W).

The Office-Home \cite{venkateswara2017deep} dataset is a larger domain adaptation benchmark containing about 15,500 images belonging to 65 different classes across 4 domains: Art (A), Clipart (C), Product (P), RealWorld (R). In addition to containing domains with a large variety of visual characteristics, the dataset presents the challenge of a large number of classes. 

\noindent \textbf{Evaluation protocol.}
We perform multiple evaluations of our model, considering at each time one of the domains as the target and the remaining ones as the source domains. We train the model until convergence on all the sources. Then, the target domain becomes available and the source domains are discarded.  At adaptation time we instantiate the domain-specific BN parameter for the target domain and perform their estimation using the newly available target images. This provides the starting point for the adaptation phase which proceeds until convergence on the target domain. In all our experiments we report the accuracy score computed on the target domain.

\noindent\textbf{Implementation details.}
We use a randomly initialized ResNet-18 as the backbone of our model. Following \cite{DBLP:journals/corr/abs-1807-06653}, we adopt an overclustering strategy that fosters the model to learn more discriminant features. Instead of using only a single head with a number of outputs equal to $C$, we add an auxiliary overclustering head with a larger number of outputs and train the two in alternating epochs. Joint training was also considered as an alternative, but performance was negatively affected. We use respectively 49, 155 and 130 output units in the auxiliary head for the PACS, Office-Home and Office31 datasets respectively. Moreover, in order to increase robustness to bad head initialization and facilitate convergence, we replicate both the standard and the overclustering head 5 times and compute the losses for the current batch on each of them, using the average loss as the optimization objective.
Further implementation details are reported in supplementary material.


\subsection{Ablation Study}
In this section, we present the results of our ablation study evaluating the impact of each of the components of \methodname. We produce different variations of our method obtained as follows: (i) Training is performed on a single source domain created by merging all the source domains; (ii) Training is performed only on a single source domain, while the others are discarded; (iii) The feature alignment via mutual information mechanism proposed in Sec.\ref{sec:domainAlignment} is removed; (iv)  Both the feature alignment via mutual information minimization mechanism and the Batch normalization feature alignment mechanism are removed, relying only on the mutual information clustering loss during training; (v) No target adaptation is performed; (vi) During adaptation the mutual information clustering loss is replaced by a prediction entropy maximization loss with threshold.

We report the quantitative results on the PACS dataset in Table \ref{table:pacsAblation}.
The ablation (iii) confirms the importance of using the mutual information loss for feature alignment during training. An analysis of the produced label assignments which we report in the supplementary material, in fact, shows that without this alignment mechanism the model produces clusters based on the domain rather than the underlying classes. The effect is that the network focuses more on learning style differences between domains rather than on semantic features, resulting in degraded performance. Removing also the BN feature alignment mechanism (iv) exacerbates the alignment problems, producing features that are not representative of the image's semantics.
Moreover, training using only a single domain as the source (ii) shows a loss in performance with respect to multi-source training, highlighting that the model acquires stronger generalization capabilities when given information about the multiple sources. Furthermore, (i) shows that it is beneficial to instantiate different BN parameters for each source domain, otherwise, the domain shift between the multiple source domains would not be mitigated.
Lastly, the proposed mutual information procedure for target adaptation outperforms the entropy-based target domain adaptation method (vi) which causes a degradation in time of the performance after a small gain in the first few epochs.

In the supplementary material we report an additional ablation on the $\alpha$ parameter introduced in Sec.\ref{sec:targetAdaptation}.

\subsection{Comparison with other methods}
\label{sec:experimentsSOTA}
We now present a comparison of \methodname~against different baseline methods. We employ two popular deep clustering methods as the first baselines, namely IIC \cite{DBLP:journals/corr/abs-1807-06653} and DeepCluster \cite{DBLP:journals/corr/abs-1807-05520}. In both cases, we train the model using only the target data. The choice of IIC is motivated by its similarity to our method and by its state-of-the-art clustering performance \cite{DBLP:journals/corr/abs-1807-06653}.
For fairness, we make use of a ResNet-18 backbone on both methods and train them on the target domain. Besides, we introduce two variations of IIC that include source information: \emph{IIC-Merge}: We train IIC on a dataset obtained by merging all the source and the target domain together; \emph{IIC+DIAL}: Following \cite{carlucci2017just}, we insert domain-specific BN layers into IIC and jointly train on all source domains plus the target domain. We also compare our method with a continuous domain adaptation strategy used in \cite{mancini2019adagraph} where we use \methodname~for training on the sources but adopt an entropy loss term for the target adaptation phase which is performed online. We also provide upper bounds for our method's performance given by SOTA domain adaptation algorithms using labeled source domains.

\begin{table}[t]
\begin{center}
\caption{Comparison of the proposed approach with SOTA on the PACS dataset. Accuracy (\%) on target domain. MS denotes multi source DA methods.}
\label{table:pacsExperiments}
\footnotesize
\begin{tabular}{lcc@{\hskip 0.075in}c@{\hskip 0.075in}c@{\hskip 0.075in}c@{\hskip 0.075in}c}
\hline\noalign{\smallskip}
 & \begin{tabular}{c}Source \\ supervision\end{tabular} & \footnotesize{C,P,S$\rightarrow$A} & \small{A,P,S$\rightarrow$C} & \small{A,C,S$\rightarrow$P} & \small{A,C,P$\rightarrow$S} & Avg\\
 
\noalign{\smallskip}
\hline
\noalign{\smallskip}

DeepCluster \cite{DBLP:journals/corr/abs-1807-05520} & $\times$ & 22.2 & 24.4 & 27.9 & 27.1 & 25.4\\
IIC \cite{DBLP:journals/corr/abs-1807-06653} & $\times$ & 39.8 & 39.6 & \textbf{70.6} & 46.6 & 49.1 \\
IIC-Merge \cite{DBLP:journals/corr/abs-1807-06653} & $\times$ & 32.2 & 33.2 & 56.4 & 30.4 & 38.1 \\
IIC \cite{DBLP:journals/corr/abs-1807-06653}~+~DIAL~\cite{carlucci2017just} & $\times$ & 30.2 & 30.5 & 50.7 & 30.7 & 35.3 \\
Continuous DA \cite{mancini2019adagraph} & $\times$ & 35.2 & 34.0 & 44.2 & 42.9 & 39.1 \\
\methodname & $\times$ & \textbf{42.1} & \textbf{44.5} & 64.4 & \textbf{51.1} & \textbf{50.5} \\
\hline\noalign{\smallskip}
\hline
\noalign{\smallskip}
AdaBN \cite{li2016revisiting} & \checkmark & $77.9$ & $74.9$ & $95.7$ & $67.7$ & $79.1$ \\
DIAL \cite{carlucci2017just}  & \checkmark & $87.3$ & $85.5$ & $97.0$ & $66.8$ & $84.2$ \\
DDiscovery \cite{mancini2018boosting} MS & \checkmark & $87.7$ & $86.9$ & $97.0$ & $69.6$ & $85.3$ \\
Jigsaw \cite{carlucci2019domain} MS & \checkmark & $84.9$ & $81.07$ & $98.0$ & $79.1$ & $85.7$ \\
AutoDIAL \cite{carlucci2017autodial} MS & \checkmark & 90.3 & 90.9 & 97.9 & 79.2 & 89.6 \\

\hline
\end{tabular}
\end{center}

\end{table}

\begin{table}[t]
\begin{center}
\caption{Comparison of the proposed approach with SOTA on the Office31 dataset. Accuracy (\%) on target domain.}
\label{table:office31Experiments}
\footnotesize
\begin{tabular}{lc@{\hskip 0.075in}c@{\hskip 0.075in}c@{\hskip 0.075in}c}
\hline\noalign{\smallskip}
 & \footnotesize{D,W$\rightarrow$A} & \small{A,W$\rightarrow$D} & \small{A,D$\rightarrow$W} & Avg\\
\noalign{\smallskip}
\hline
\noalign{\smallskip}

DeepCluster \cite{DBLP:journals/corr/abs-1807-05520} & 19.6 & 18.7 & 18.9 & 19.1 \\
IIC \cite{DBLP:journals/corr/abs-1807-06653} & 31.9 & 34.0 & 37.0 & 34.3 \\
IIC-Merge \cite{DBLP:journals/corr/abs-1807-06653} & 29.1 & \textbf{36.1} & 33.5 & 32.9 \\
IIC \cite{DBLP:journals/corr/abs-1807-06653}~+~DIAL~\cite{carlucci2017just} & 28.1 & 35.3 & 30.9 & 31.4 \\
Continuous DA \cite{mancini2019adagraph} & 20.5 & 28.8 & 30.6 & 26.6 \\
\methodname & \textbf{33.4} & \textbf{36.1} & \textbf{37.5} & \textbf{35.6} \\

\hline
\end{tabular}
\end{center}

\end{table}

\begin{table}[t]
\begin{center}
\caption{Comparison of the proposed approach with SOTA on the Office-Home dataset. Accuracy (\%) on target domain.}
\label{table:officeHomeExperiments}
\footnotesize
\begin{tabular}{lc@{\hskip 0.075in}c@{\hskip 0.075in}c@{\hskip 0.075in}c@{\hskip 0.075in}c}
\hline\noalign{\smallskip}
 & \footnotesize{C,P,R$\rightarrow$A} & \small{A,P,R$\rightarrow$C} & \small{A,C,R$\rightarrow$P} & \small{A,C,P$\rightarrow$R} & Avg\\
\noalign{\smallskip}
\hline
\noalign{\smallskip}

DeepCluster \cite{DBLP:journals/corr/abs-1807-05520} & 8.9 & 11.1 & 16.9 & 13.3 & 12.6 \\
IIC \cite{DBLP:journals/corr/abs-1807-06653} & \textbf{12.0} & 15.2 & 22.5 & \textbf{15.9} & 16.4 \\
IIC-Merge \cite{DBLP:journals/corr/abs-1807-06653} & 11.3 & 13.1 & 16.2 & 12.4 & 13.3 \\
IIC \cite{DBLP:journals/corr/abs-1807-06653}~+~DIAL~\cite{carlucci2017just} & 10.9 & 12.9 & 15.4 & 12.8 & 13.0 \\
Continuous DA \cite{mancini2019adagraph} & 10.2 & 11.5 & 13.0 & 11.7 & 11.6 \\
\methodname & \textbf{12.0} & \textbf{16.2} & \textbf{23.9} & 15.7 & \textbf{17.0} \\

\hline
\end{tabular}
\end{center}

\end{table}

We report the performance of our method on the PACS dataset in Table~\ref{table:pacsExperiments}. Our method performs substantially better than the DeepCluster and IIC baselines in the Art, Cartoon and Sketch domains with accuracy gains in the range from 2.3\% to 4.9\% with respect to IIC, while on the Photo domain our approach does not reach its performance. Moreover, our adaptation procedure outperforms the continuous domain adaptation baseline whose entropy loss does capture the semantic aspects given by our mutual information approach. The comparison with the upper bounds shows instead the obvious advantage of using supervision on the source domains. Due to the large difference of this setting with the proposed one, we omit these upper bounds from the successive evaluations.

In Table \ref{table:office31Experiments} we report the performance of \methodname~on the Office31 dataset. The proposed approach achieves state of the art results, performing better than both DeepCluster and IIC on all domains with accuracy gains from 0.5\% to 2.1\% with respect to the strongest baseline.

Lastly, Table \ref{table:officeHomeExperiments} shows the results obtained on the Office-Home dataset. The proposed approach performs significantly better than the DeepCluster baseline on each domain and performs better than IIC on the Clipart and Product domains. Similarly to the results on the PACS dataset, our target adaptation procedure performs better than the continuous domain adaptation strategy.

\subsection{Limited target data scenario}

\begin{figure}[t]
\centering
\includegraphics[width=0.99\textwidth]{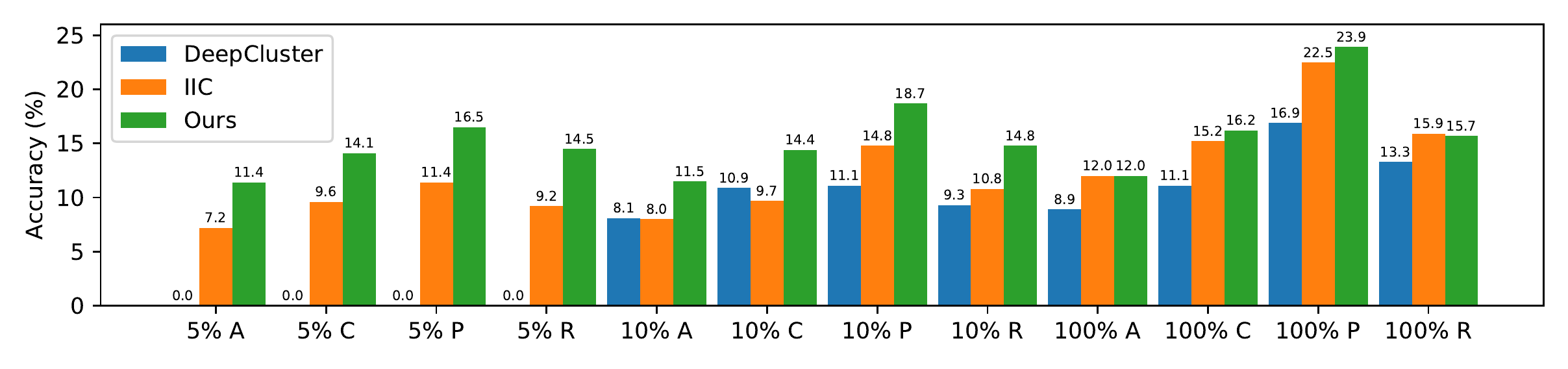}

\caption{Comparison of the proposed approach with SOTA on the Office-Home dataset in the limited target data scenario. Labels express the target domain (A,C,P or R) and the percentage of images used in the target domain.} 

\label{fig:officeHomeExperimentsFew}
\end{figure}

One of the major advantages of \methodname~is the possibility of extracting semantic features from the source domains that directly transfer to the target domain. This makes it particularly suitable for the task of domain adaptation when few target samples are available. We repeat the same experiments of Sec.\ref{sec:experimentsSOTA} on the Office-Home dataset where the source domains are not altered and we consider a target domain built by randomly sampling a given portion of images in each class of the original target domain. We show the numerical results in Fig.\ref{fig:officeHomeExperimentsFew}. We achieve a large performance boost compared to the baselines, in particular, we achieve an average 4.1\% and 4.9\% increase in accuracy with respect to IIC when 10\% and 5\% of the target images are available. Note that DeepCluster is not able to operate in the 5\% scenario due to an insufficient number of target samples.

\section{Conclusions}

In this paper, we propose a novel domain adaptation setting and show it is possible to transfer knowledge from multiple source domains to a target domain when both sources and target data have no annotations. Our method makes use of a novel information-theoretic loss for feature alignment and couples it with domain-alignment layers to discover semantic labels from the source domains. When target data becomes available, we perform adaptation without requiring the availability of source data.
We achieve state-of-the-art performance on three widely used domain adaptation datasets and show a clear advantage of the proposed approach under low target data conditions. Future works will consider the adaptation of the approach to the unsupervised segmentation scenario.

\section*{Acknowledgements}
We acknowledge financial support from H2020 EU project SPRING - Socially Pertinent Robots in Gerontological Healthcare. This work was carried out under the ``Vision and Learning joint Laboratory'' between FBK and UNITN.

%
%
\bibliographystyle{splncs04}
\bibliography{egbib}
\end{document}


\pagestyle{headings}
\mainmatter
\def\ECCVSubNumber{6436}  
\newcommand{\steph}[1]{\textcolor{red}{Steph: #1}}
\newcommand{\eli}[1]{{\color{magenta}#1}}
\newcommand{\willi}[1]{\textcolor{blue}{#1}}
\newcommand{\methodname}{ACIDS}

\title{Learning to Cluster under Domain Shift: Supplementary Material} 



\titlerunning{Learning to Cluster under Domain Shift}
%
\author{Willi Menapace\inst{1} \and 
St\'{e}phane Lathuili\`{e}re\inst{3} 
\and
Elisa Ricci\inst{1,2}}
\authorrunning{W. Menapace et al.}
%
\institute{University of Trento, Trento, Italy \and
Fondazione Bruno Kessler, Trento, Italy
\and
LTCI, T\'{e}l\'{e}com Paris, Institut Polytechnique de Paris, Palaiseau, France\\
\email{willi.menapace@gmail.com}\\
}

\maketitle

\section{Additional Implementation Details}
We use a ResNet-18 backbone for all the experiments and report the hyperparameters used for each in Table~\ref{table:hyperparameters}. For all datasets we compose $\mathcal{T}$ using random crops, random horizontal flips and random hue, saturation and brightness changes. The input resolution to the network is $64\!\times\!64$px and on our GPUS with 8GB of memory allows the use of a maximum batch size of 162 images.
The value of $\alpha$ is a function of the number of ground truth classes and the number of source domains in each dataset. A larger number of ground truth classes $C$ causes a larger probability matrix $P_{cc'}$ to be estimated, while a higher number of source domains empirically causes more instability, probably due to a higher variance of features despite alignment. In particular, the PACS \cite{li2017deeper} dataset with $C=7$ does not suffer much from training instability problems, so a value of $\alpha=0.7$ is chosen. On the other hand, the Office31 \cite{saenko2010adapting} and the Office-Home \cite{venkateswara2017deep} datasets with respectively $C=31$ and $C=65$ pose more stability problems. As we note in our Ablation Study on the main paper, a value of $\alpha=0.1$ produces the best results on the Office31 dataset when training on all the dataset domains. Since in the standard experimental setting, however, we consider one domain as the target and train on one less source domain, we decide to use a higher value of $\alpha=0.2$.
The number of classes produced by the overclustering head $C_{oc}$ is chosen following \cite{DBLP:journals/corr/abs-1807-06653} which obtains the best results when choosing a value from 5 to 7 times $C$. Note that on the Office-Home dataset, due to the high number of ground truth classes, a factor of 2 is used.
We use the Adam optimizer with learning rate $10^{-4}$ in all the experiments.

\begin{table}
\begin{center}
\caption{Hyperparameter values used during the experiments. $s$ denotes the number of times each head is replicated to improve training stability, $C_{oc}$ represents the number of output classes used in the auxiliary overclustering head.}
\label{table:hyperparameters}
\begin{tabular}{llccccc}
\hline\noalign{\smallskip}
Dataset & Task & BS & $\alpha$ & s & $C$ & $C_{oc}$\\
\noalign{\smallskip}
\hline
\noalign{\smallskip}

PACS & \footnotesize{C,P,S$\rightarrow$A} & 162 & 0.7 & 5 & 7 & 49 \\
PACS & \footnotesize{A,P,S$\rightarrow$C} & 162 & 0.7 & 5 & 7 & 49 \\
PACS & \footnotesize{A,C,S$\rightarrow$P} & 162 & 0.7 & 5 & 7 & 49 \\
PACS & \footnotesize{A,C,P$\rightarrow$S} & 162 & 0.7 & 5 & 7 & 49 \\

\hline
\noalign{\smallskip}

Office31 & \footnotesize{D,W$\rightarrow$A} & 162 & 0.2 & 5 & 31 & 155 \\
Office31 & \footnotesize{A,W$\rightarrow$D} & 162 & 0.2 & 5 & 31 & 155 \\
Office31 & \footnotesize{A,D$\rightarrow$W} & 162 & 0.2 & 5 & 31 & 155 \\

\hline
\noalign{\smallskip}

Office-Home & \footnotesize{C,P,R$\rightarrow$A} & 162 & 0.2 & 5 & 65 & 130 \\
Office-Home & \footnotesize{A,P,R$\rightarrow$C} & 162 & 0.2 & 5 & 65 & 130 \\
Office-Home & \footnotesize{A,C,R$\rightarrow$P} & 162 & 0.2 & 5 & 65 & 130 \\
Office-Home & \footnotesize{A,C,P$\rightarrow$R} & 162 & 0.2 & 5 & 65 & 130 \\

\hline
\end{tabular}
\end{center}

\end{table}

\section{Additional Details about Baselines}
As stated in the main paper, we make use of the IIC~\cite{DBLP:journals/corr/abs-1807-06653} and DeepCluster~\cite{DBLP:journals/corr/abs-1807-05520} algorithms as baselines for the evaluation of our method.
With regards to the IIC baseline, we make use of the same ResNet-18 backbone and the same hyperparameters reported in Table~\ref{table:hyperparameters} to guarantee fairness in the evaluation. Note that we do not employ the $\alpha$-smoothing strategy in the IIC baseline in order to follow their exact implementation.
For DeepCluster, we use a ResNet-18 backbone and train it using an SGD optimizer with learning rate $10^{-2}$ for all the experiments. Despite \cite{DBLP:journals/corr/abs-1807-05520} suggests the use of a number of clusters for self-supervision during training equal to 10 times the number of ground truth classes, we employed smaller factors due to the small number of samples in the datasets which does not allow the intermediate K-means procedure to work effectively with a high number of clusters.

\section{Additional Target Adaptation Experimental Results}
\begin{figure}[t]
\hspace{-0.2cm}
\centering
\hspace*{\fill}
\begin{minipage}[b]{0.49\textwidth}
    \begin{tabular}{c}
         \includegraphics[width=\textwidth]{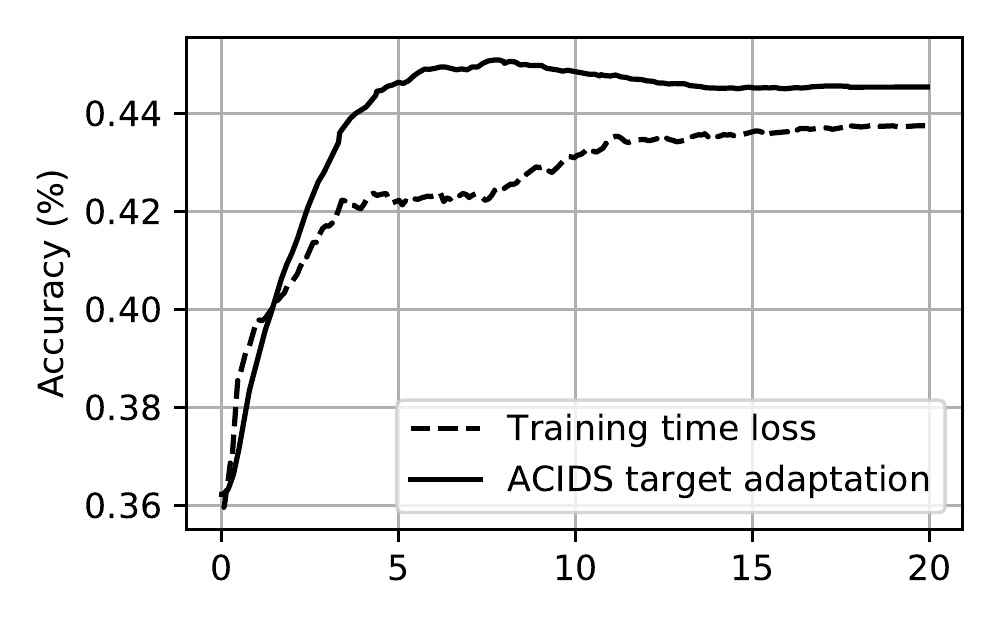}  \\
         \raisebox{1.0\height}{~~~~~~~~Cartoon}
    \end{tabular}
    
  \end{minipage}
  \hfill
  \begin{minipage}[b]{0.49\textwidth}
  \begin{tabular}{c}
         \includegraphics[width=\textwidth]{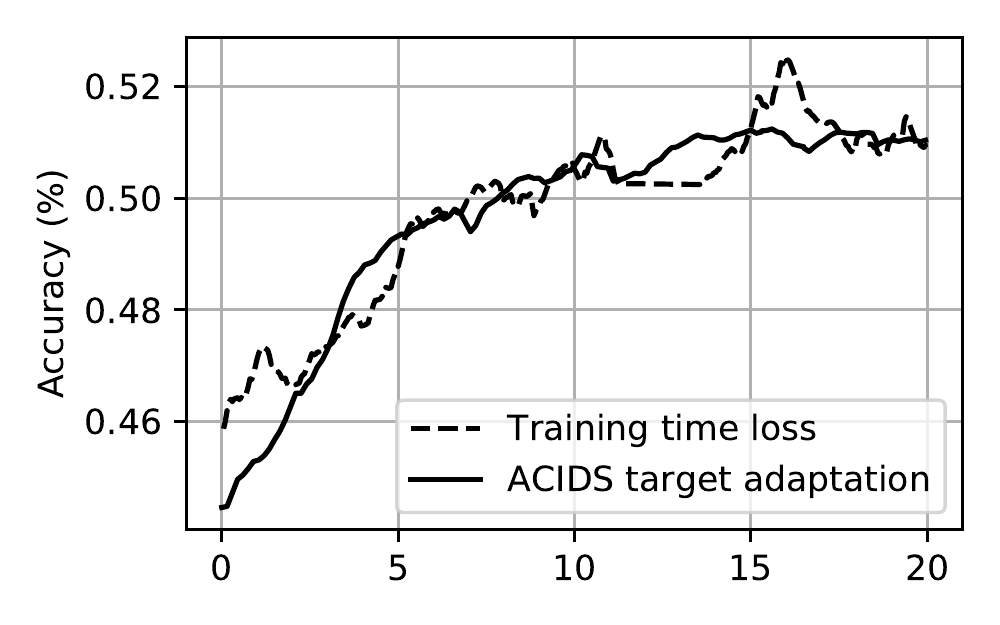}  \\
         \raisebox{1.0\height}{~~~~~~~~~Sketch}
    \end{tabular}
  \end{minipage}
\hspace*{\fill}

\caption{Evolution of accuracy on the target domain during the target adaptation phase using Cartoon (left) and Sketch (right) as target domains. The solid line refers to the ACIDS target adaptation procedure, the dashed one refers to adaptation using the same mutual information maximization procedure used during training. Time on the x-axis is expressed in thousands of optimization steps.}

\label{fig:trainingCurves}
\end{figure}

In this section, we propose to further investigate the effectiveness of the ACIDS target domain adaptation procedure. We perform training on the source domains and, starting from the same network parameters, we perform two different target adaptation procedures. The first uses the ACIDS adaptation procedure described in the main paper, the second performs adaptation using the same mutual information loss used at training time, computed on the target domain. In order to illustrate the stable convergence of the proposed model, we show in Fig.\ref{fig:trainingCurves} the evolution of the accuracy on the target domain while performing adaptation on two domains of the PACS dataset. The proposed target adaptation procedure leads to faster convergence and higher accuracy on the Cartoon domain (Fig.\ref{fig:trainingCurves}-left), while it produces no appreciable effects on the Sketch domain (Fig.\ref{fig:trainingCurves}-right).

\section{Additional Parameter Ablation}

We perform an evaluation of the effect of the $\alpha$ parameter described in Sec.3.3 on the main paper. Since it would be computationally expensive to run separate training processes for every value of $\alpha$ and every domain, we adopt an evaluation protocol where we perform training without target adaptation, considering every domain as a source domain. For these experiments, we choose the Office31 dataset that is especially challenging in terms of optimization because of its high number of classes $C$.

\begin{figure}[t]
\centering
\includegraphics[width=0.99\textwidth]{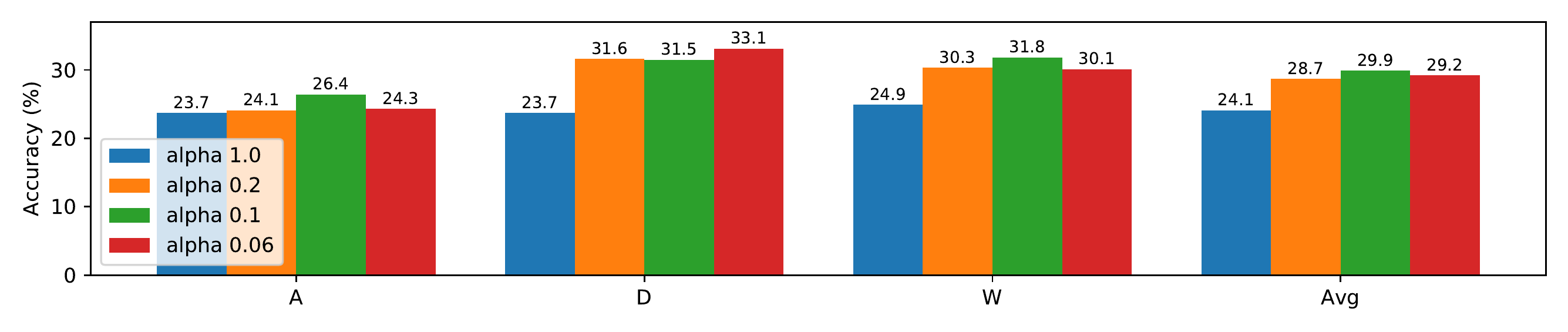}

\caption{Ablation of the $\alpha$ parameter on the Office 31 dataset. Labels express the source domain (A, D or W). Results expressed in accuracy (\%).}

\label{fig:alphaAblation}
\end{figure}

Fig.\ref{fig:alphaAblation} reports the numerical evaluation results. Without using our stabilization method ($\alpha=1.0$), we obtain degraded results due to noise in the estimation of $P_{cc'}$. Lowering the value to $\alpha=0.1$ improves the estimation, achieving +5.8\% average accuracy with respect to $\alpha=0$. When further decreasing the value, however, accuracy starts to decrease. The estimation, in this case, becomes incorrect because it is influenced by network parameters that differ too much from the current ones.

\section{Feature Alignment via Mutual Information Qualitative Evaluation}

\begin{figure}[t]
\hspace{-0.2cm}
\centering
\hspace*{\fill}
\begin{minipage}[b]{0.25\textwidth}
    \includegraphics[trim=70 40 120 45,clip,width=\textwidth]{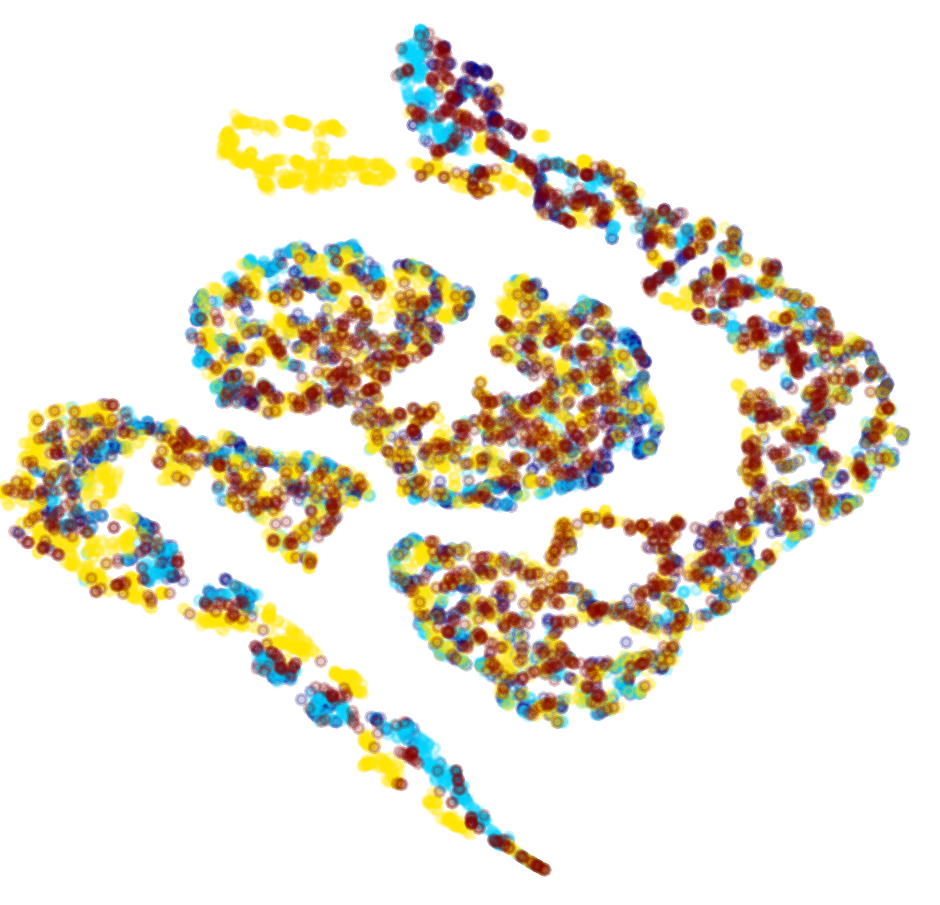}
  \end{minipage}
  \hfill
  \begin{minipage}[b]{0.25\textwidth}
    \includegraphics[trim=70 40 120 45,clip,width=\textwidth]{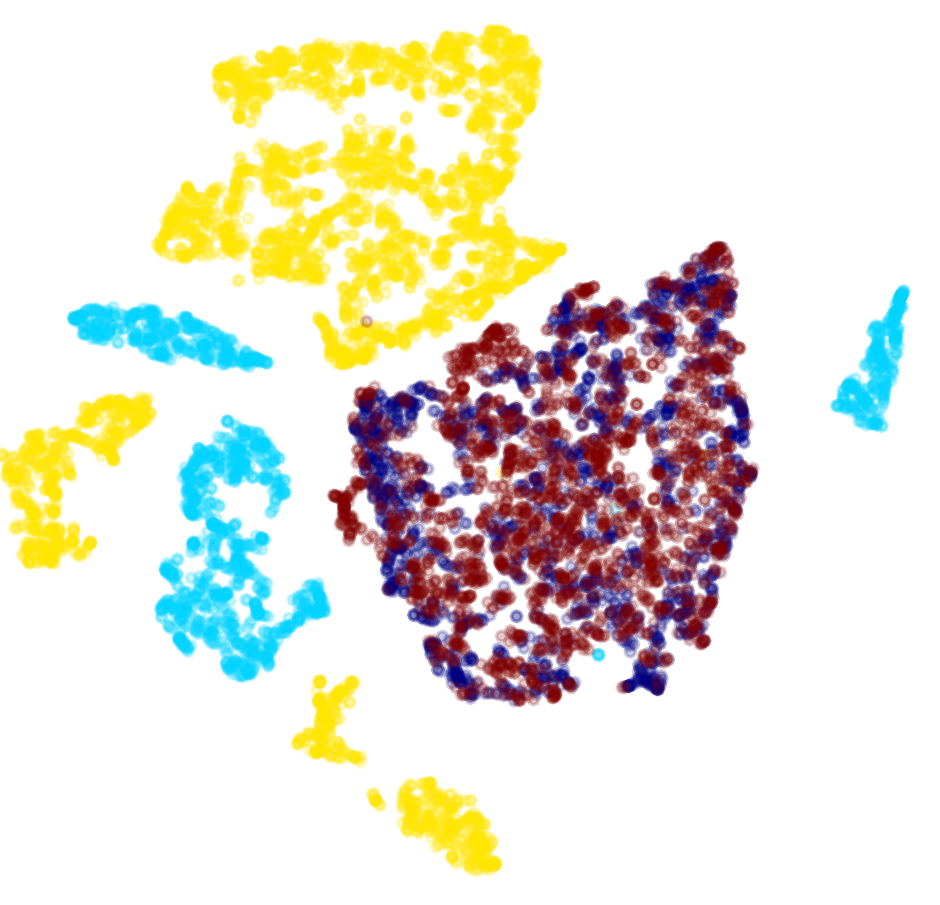}
  \end{minipage}
\hspace*{\fill}

\caption{t-SNE visualizations of the feature space extracted before the classification heads for the proposed method (left) and the same method without the mutual information loss for feature alignment Eq.(4) on the main paper (right), using PACS Cartoon as the target domain. Colors represent the different target domains. While in the proposed method the distributions of the domains align (left), when the feature alignment loss is removed (right) the method produces clusters based on the style rather than the content information (best viewed in color).}
\label{fig:tsneSup}
\end{figure}

In Fig.\ref{fig:tsneSup} we report a qualitative analysis of the effect of the proposed mutual information minimization procedure for feature alignment on the feature space. The analysis shows that without the proposed procedure, the method produces clusters based on style rather than the image semantics, while the desired domain alignment is obtained when employing the proposed method.

\section{Qualitative Clustering Results}
\begin{figure}
\centering
\hspace*{\fill}
\begin{minipage}[b]{0.24\textwidth}
    \hspace{-0.1cm}
    \includegraphics[width=\textwidth]{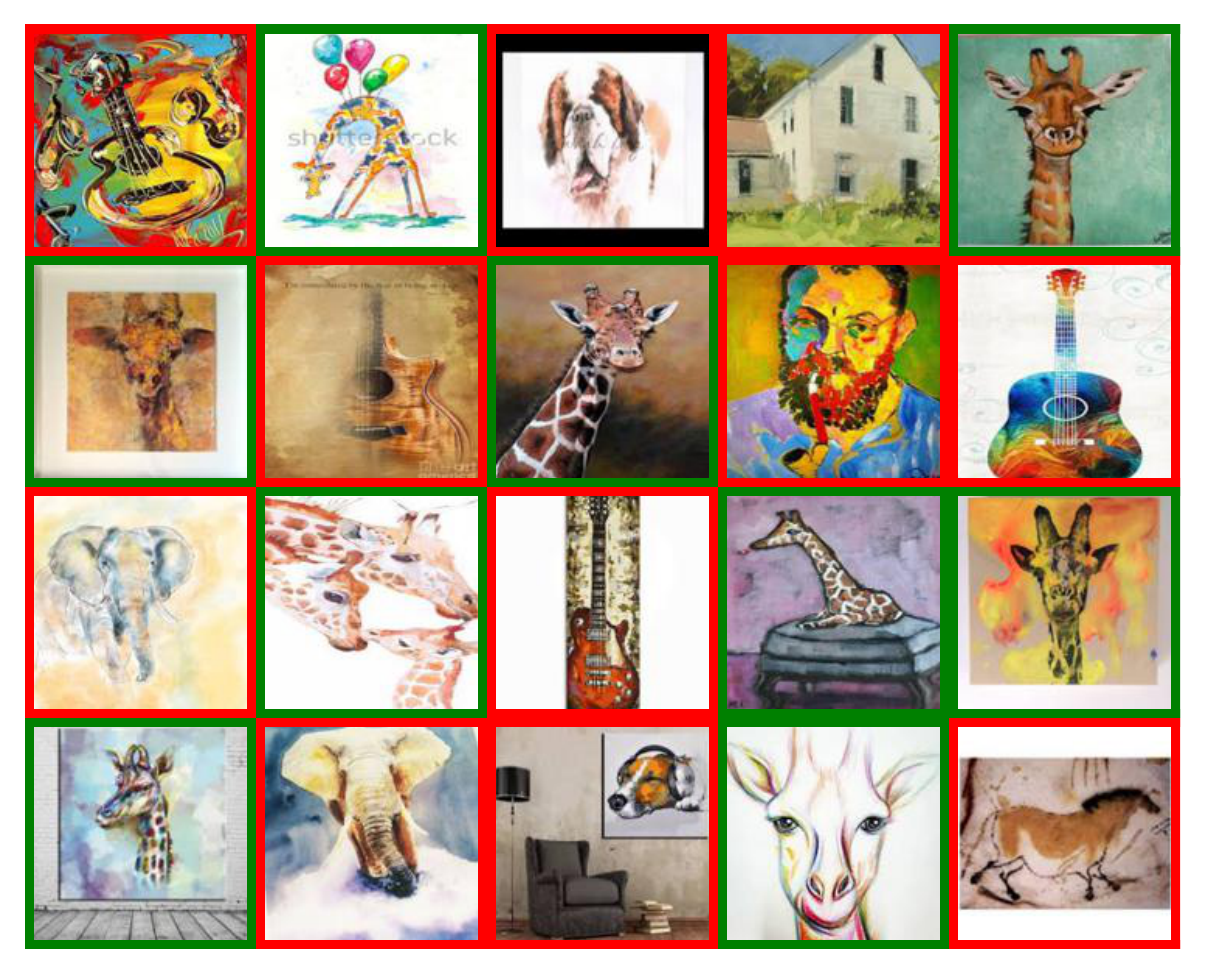}
  \end{minipage}
  \hfill
  \begin{minipage}[b]{0.24\textwidth}
    \includegraphics[width=\textwidth]{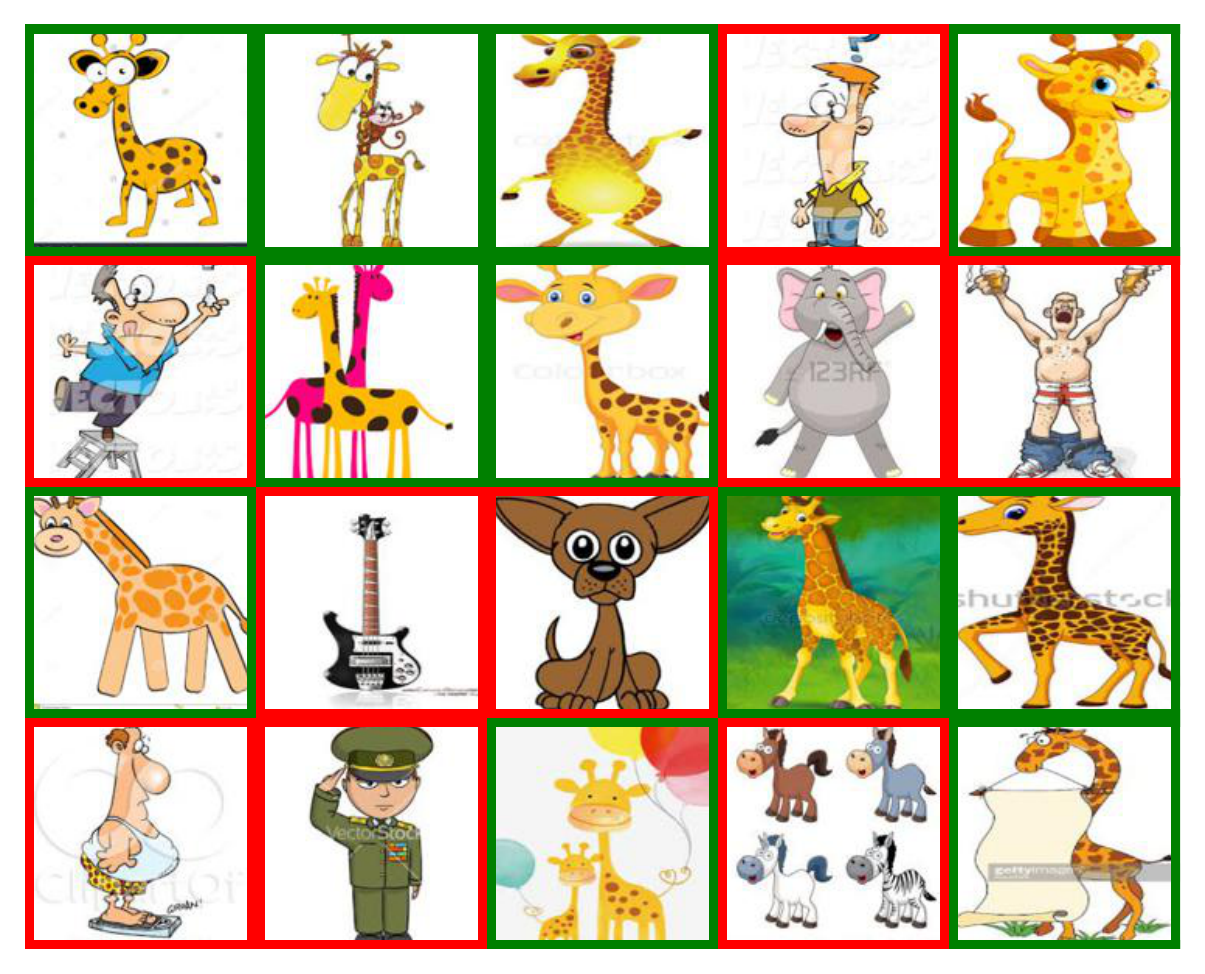}
  \end{minipage}
  \hfill
  \begin{minipage}[b]{0.24\textwidth}
    \includegraphics[width=\textwidth]{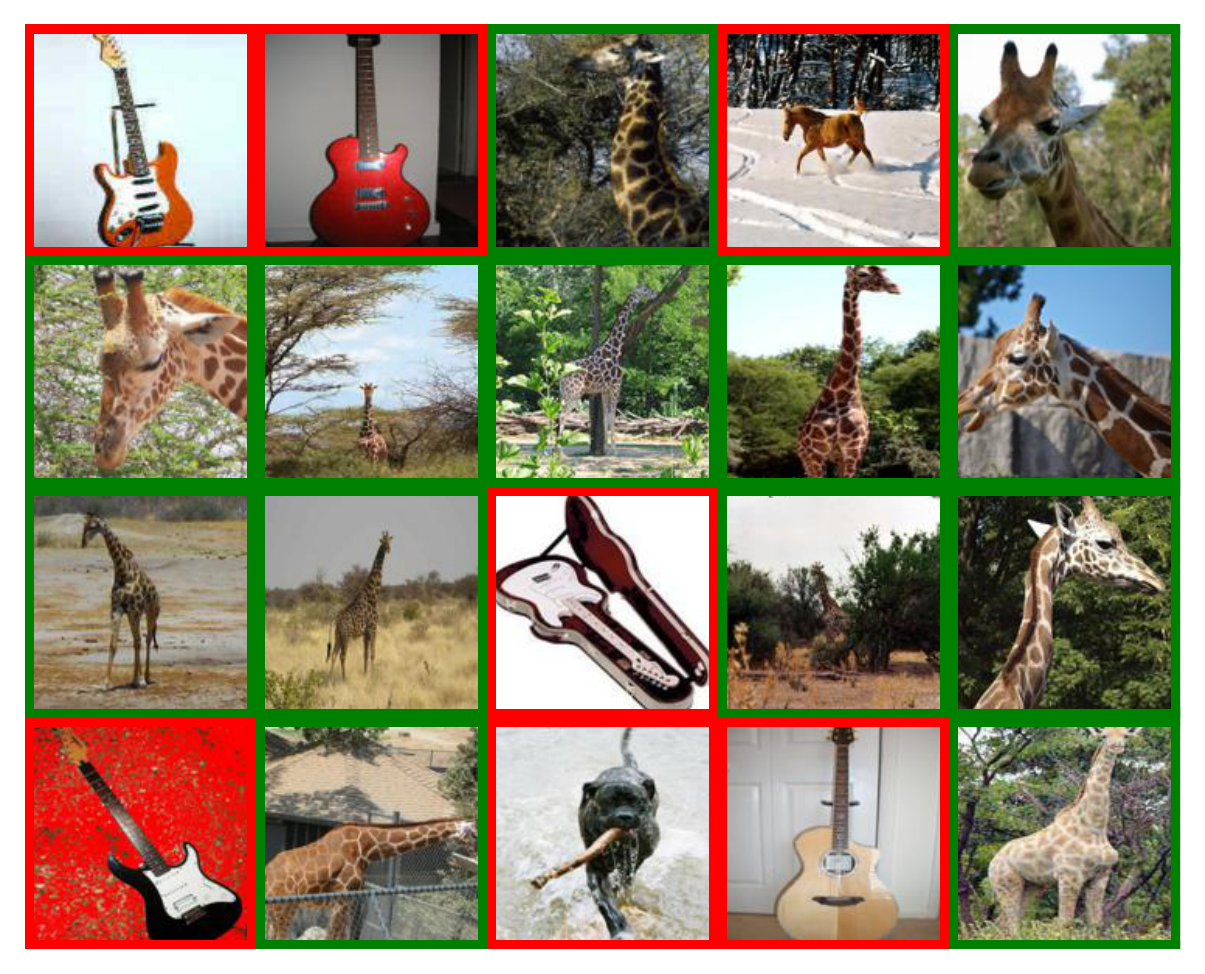}
  \end{minipage}
  \hfill
  \begin{minipage}[b]{0.24\textwidth}
    \includegraphics[width=\textwidth]{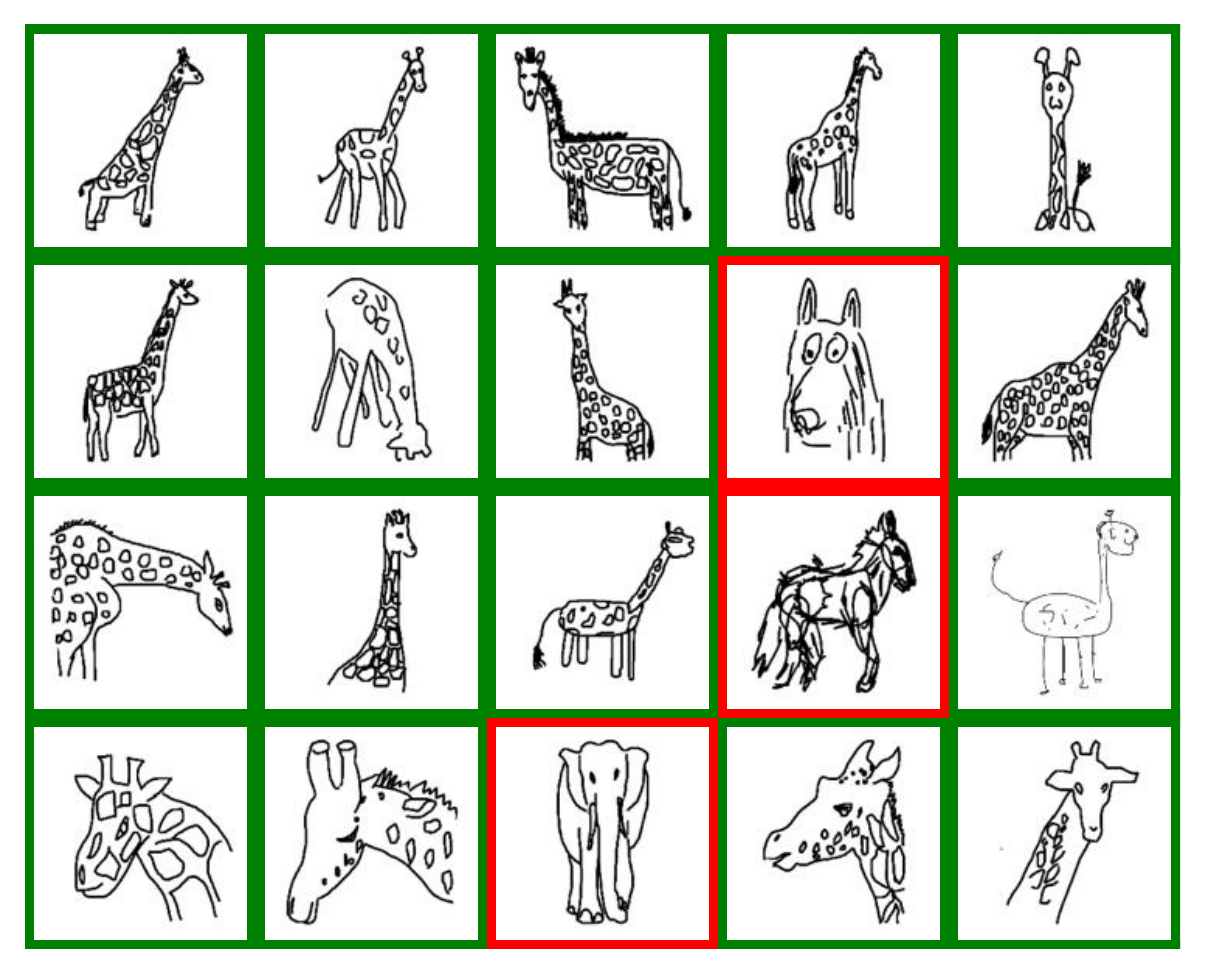}
  \end{minipage}
\hspace*{\fill}

\caption{Clusters corresponding to the PACS ``Giraffe'' class from Art, Cartoon, Photo and Sketch domains.}

\label{fig:pacsGiraffe}
\end{figure}

\begin{figure}
\centering
\hspace*{\fill}
\begin{minipage}[b]{0.24\textwidth}
    \hspace{-0.1cm}
    \includegraphics[width=\textwidth]{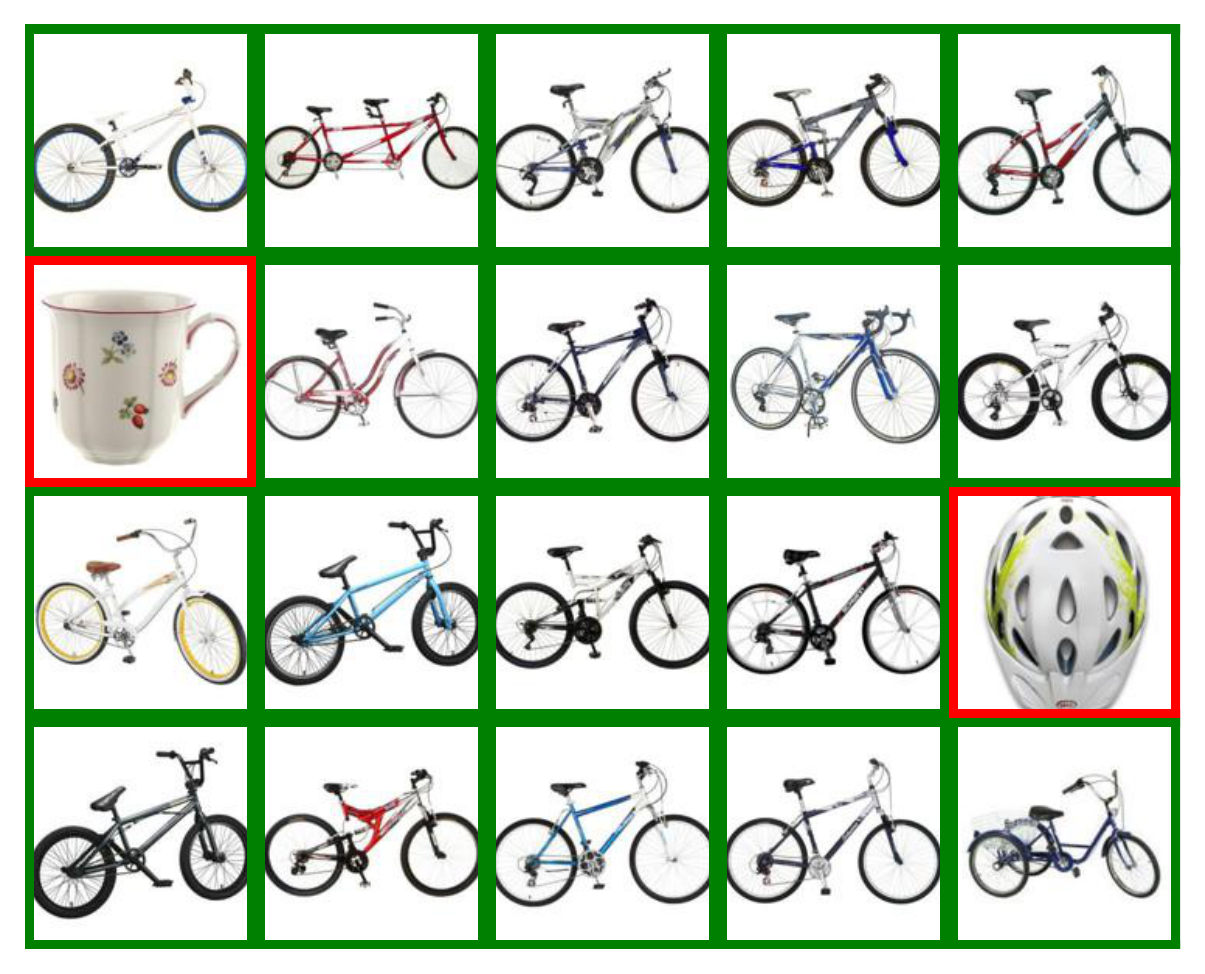}
  \end{minipage}
  \hfill
  \begin{minipage}[b]{0.24\textwidth}
    \includegraphics[width=\textwidth]{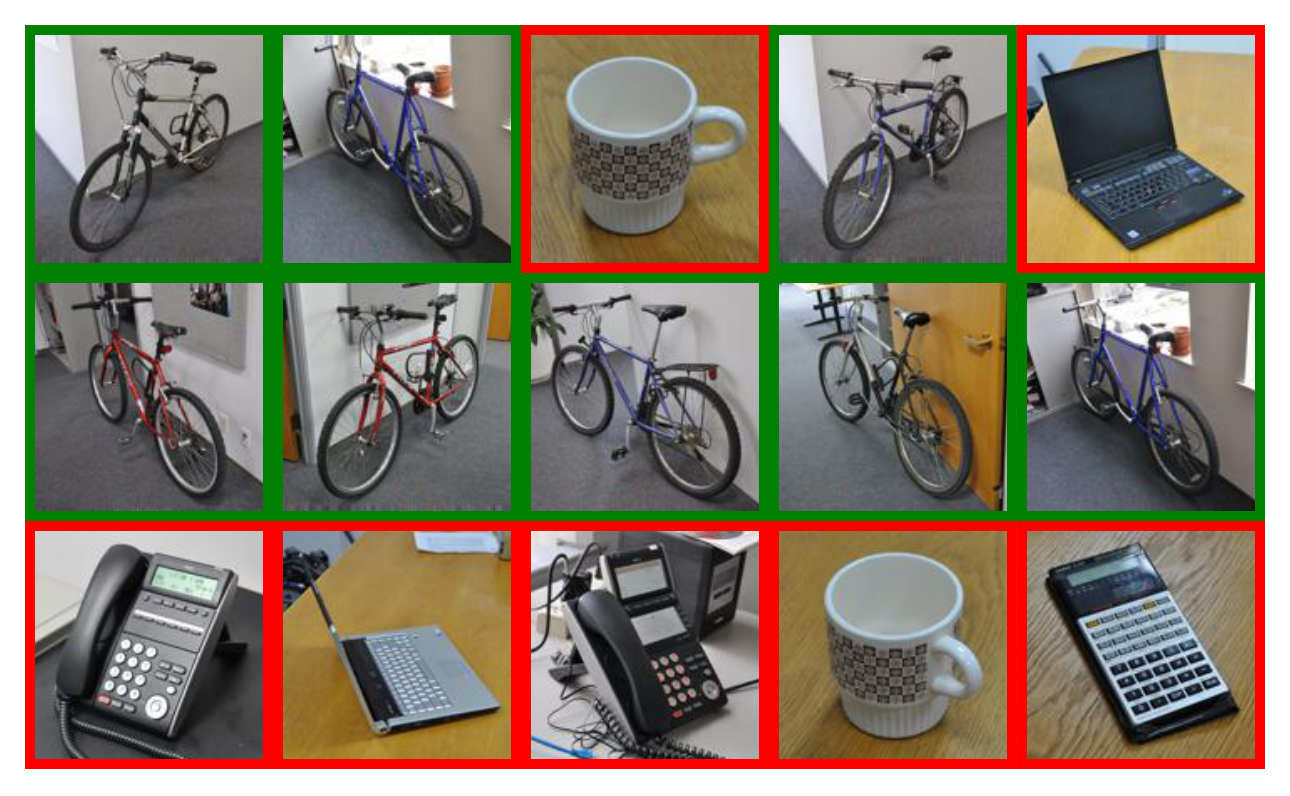}
  \end{minipage}
  \hfill
  \begin{minipage}[b]{0.24\textwidth}
    \includegraphics[width=\textwidth]{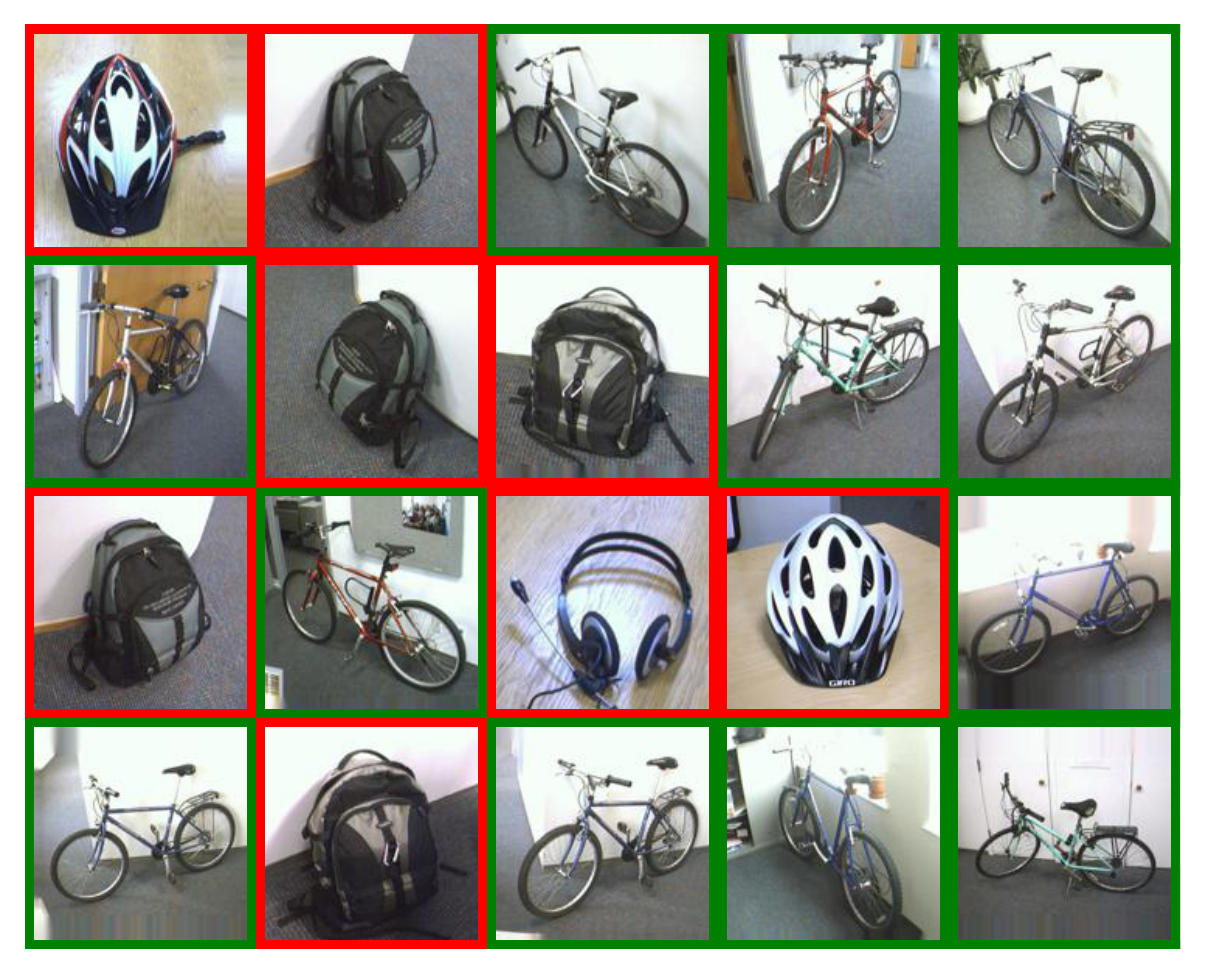}
  \end{minipage}
\hspace*{\fill}

\caption{Clusters corresponding to the Office31 ``Bike'' class from Amazon, DSLR and Webcam domains.}
\label{fig:office31Bike}

\label{fig:tsne}
\end{figure}

\begin{figure}
\centering
\hspace*{\fill}
\begin{minipage}[b]{0.24\textwidth}
    \hspace{-0.1cm}
    \includegraphics[width=\textwidth]{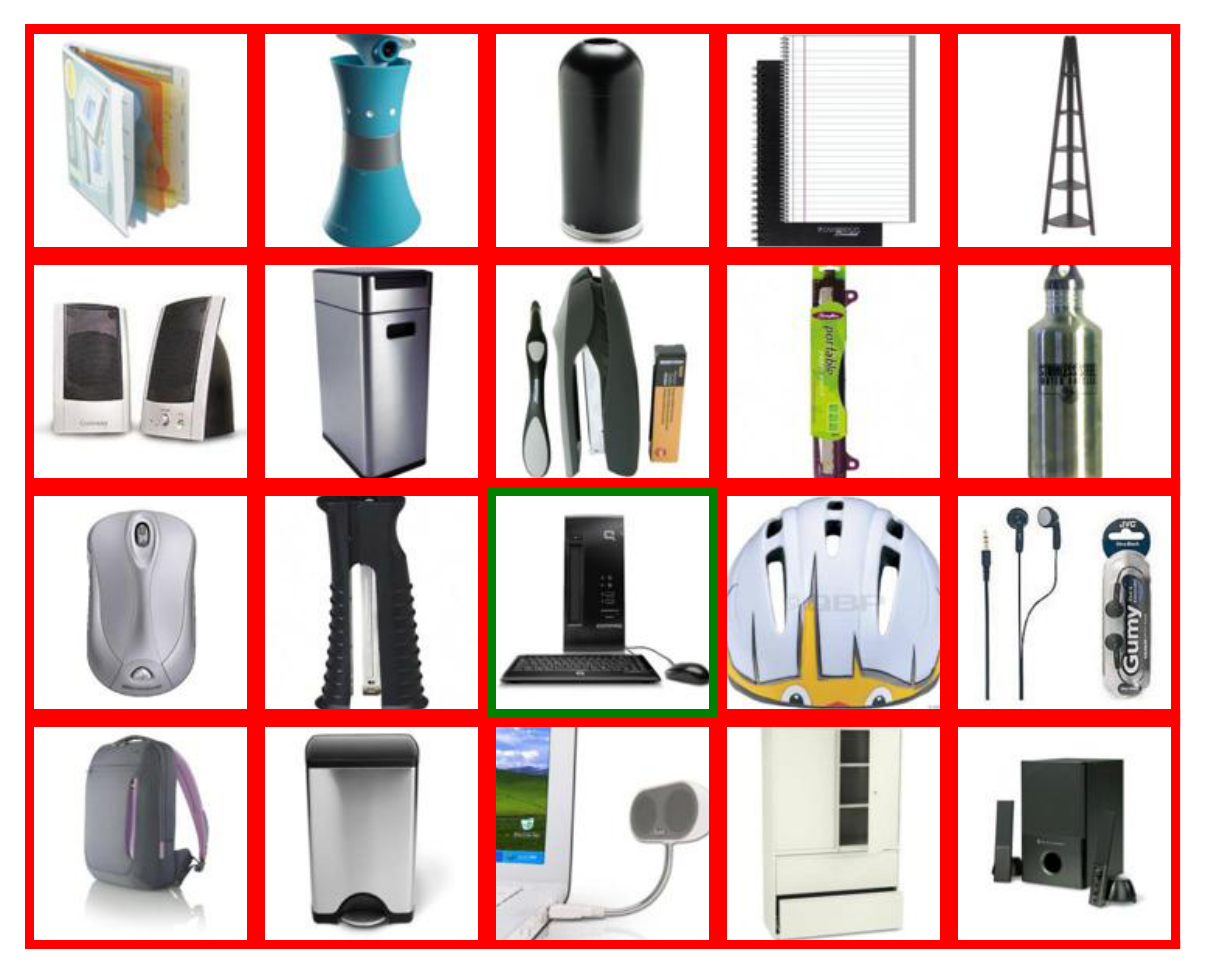}
  \end{minipage}
  \hfill
  \begin{minipage}[b]{0.24\textwidth}
    \includegraphics[width=\textwidth]{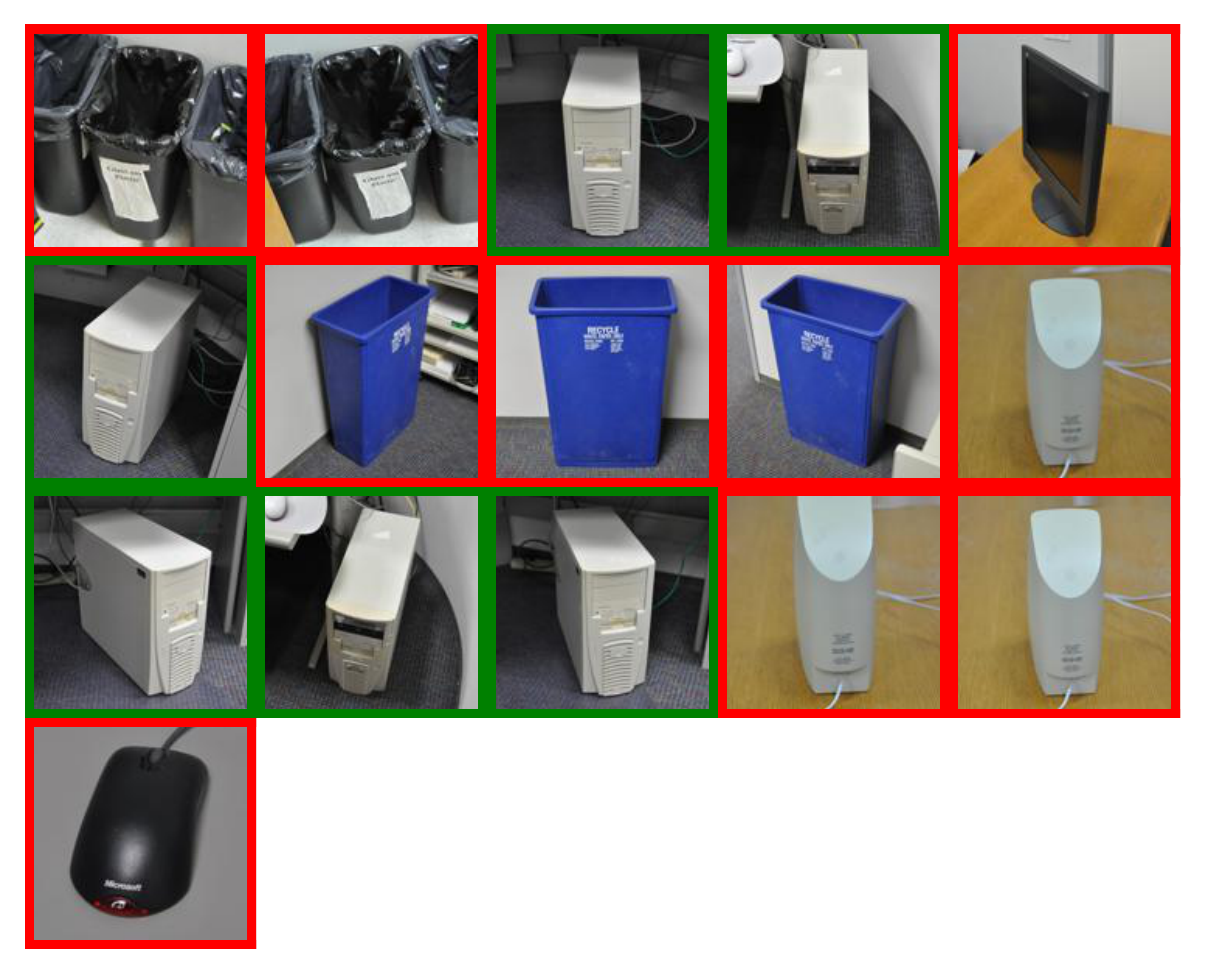}
  \end{minipage}
  \hfill
  \begin{minipage}[b]{0.24\textwidth}
    \includegraphics[width=\textwidth]{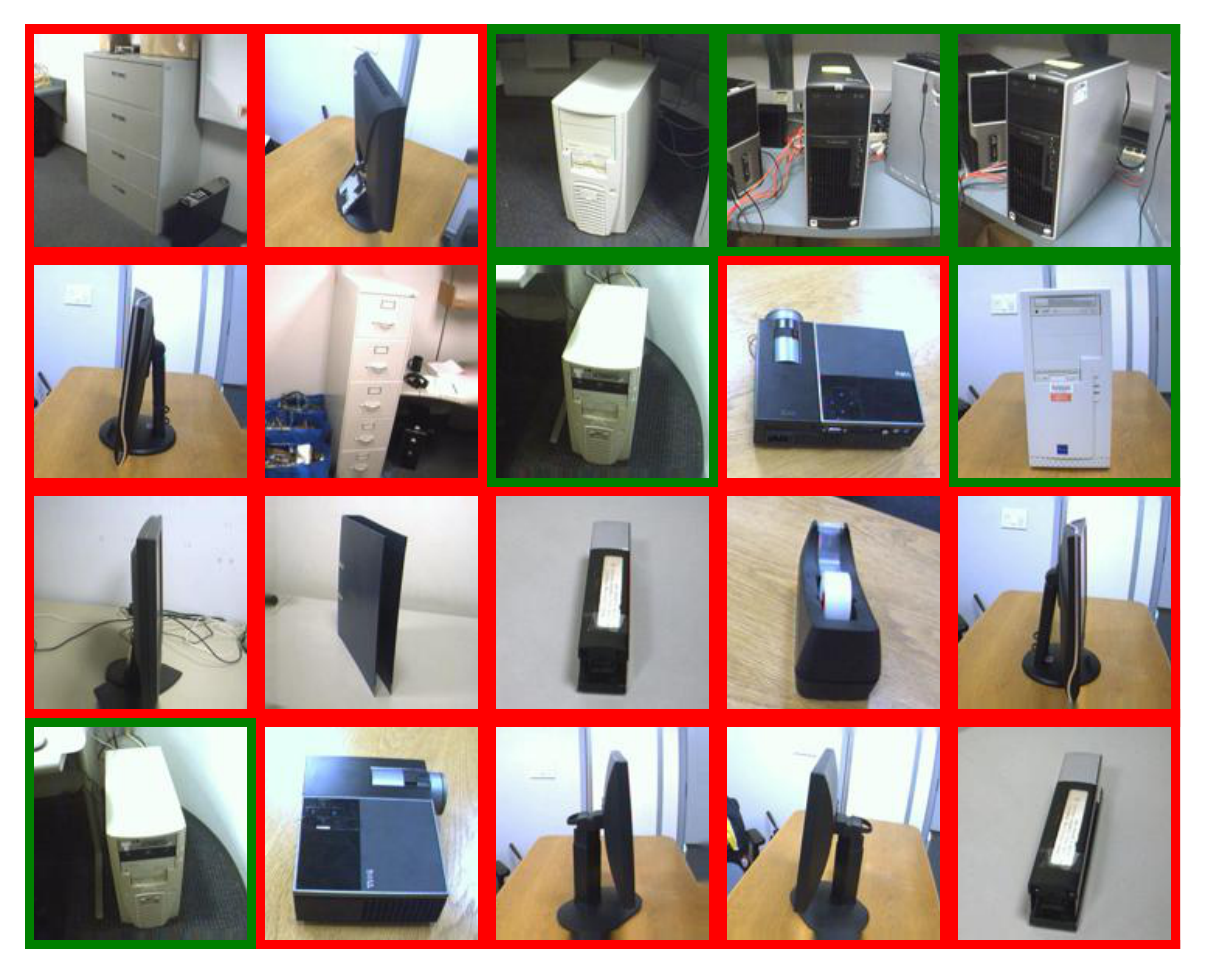}
  \end{minipage}
\hspace*{\fill}

\caption{Clusters corresponding to the Office31 ``Desktop Computer'' class from Amazon, DSLR and Webcam domains.}
\label{fig:office31DesktopComputer}

\label{fig:tsne}
\end{figure}

In this section, we present qualitative clustering results produced by our method. Each cluster visualization corresponds to the results produced by the model after adaptation to the corresponding target domain.
A visual inspection of the produced clusters reveals that classes with the most distinctive features such as ``Giraffe'' in PACS (Fig.\ref{fig:pacsGiraffe}) or ``Bike'' in Office31 (Fig.\ref{fig:office31Bike}) tend to be clustered best, while classes with shapes similar to others tend to be confused like the ``Desktop Computer'' class in Office31. (Fig.\ref{fig:office31DesktopComputer}).
\begin{figure}
\centering
\hspace*{\fill}
\begin{minipage}[b]{0.24\textwidth}
    \hspace{-0.1cm}
    \includegraphics[width=\textwidth]{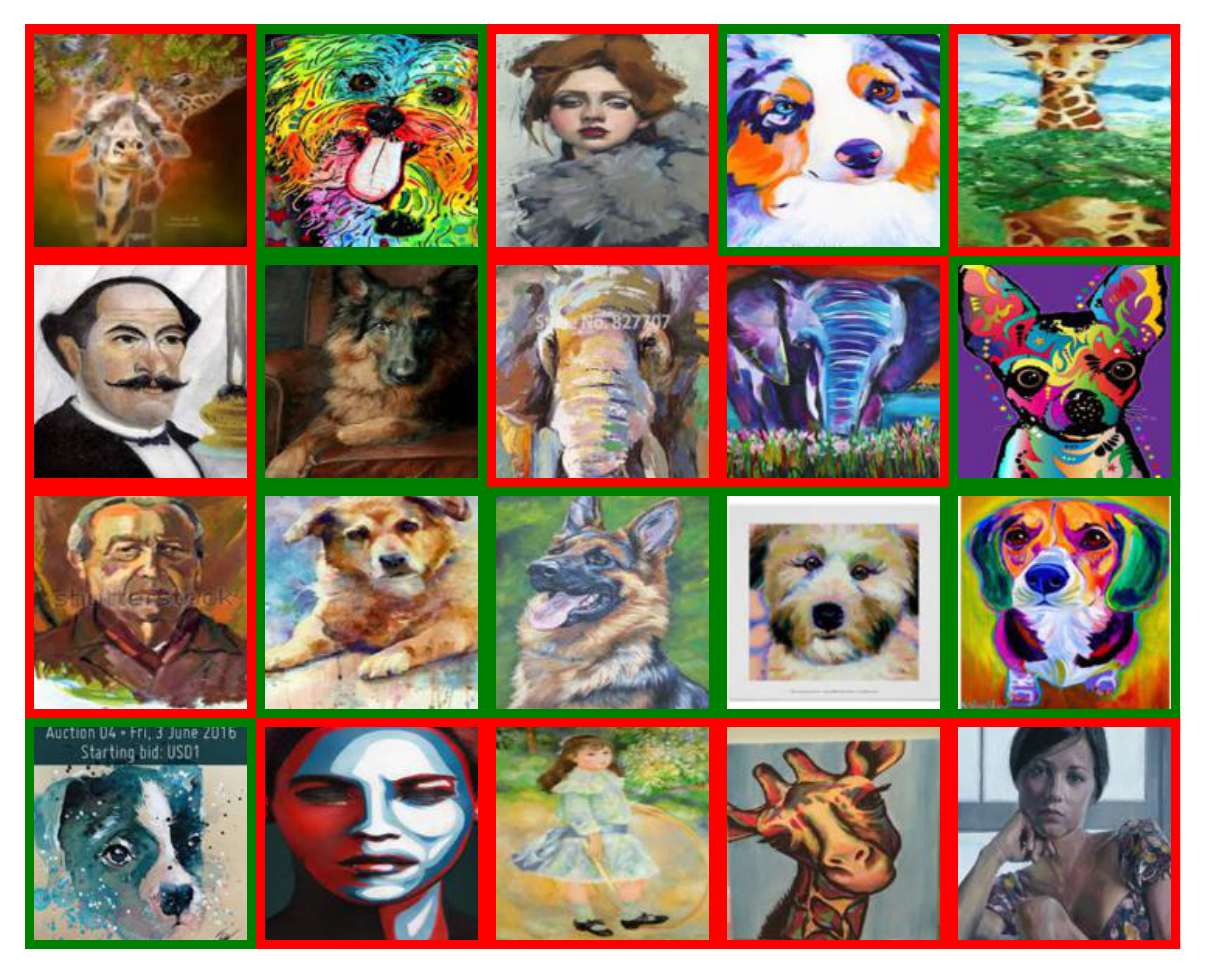}
  \end{minipage}
  \hfill
  \begin{minipage}[b]{0.24\textwidth}
    \includegraphics[width=\textwidth]{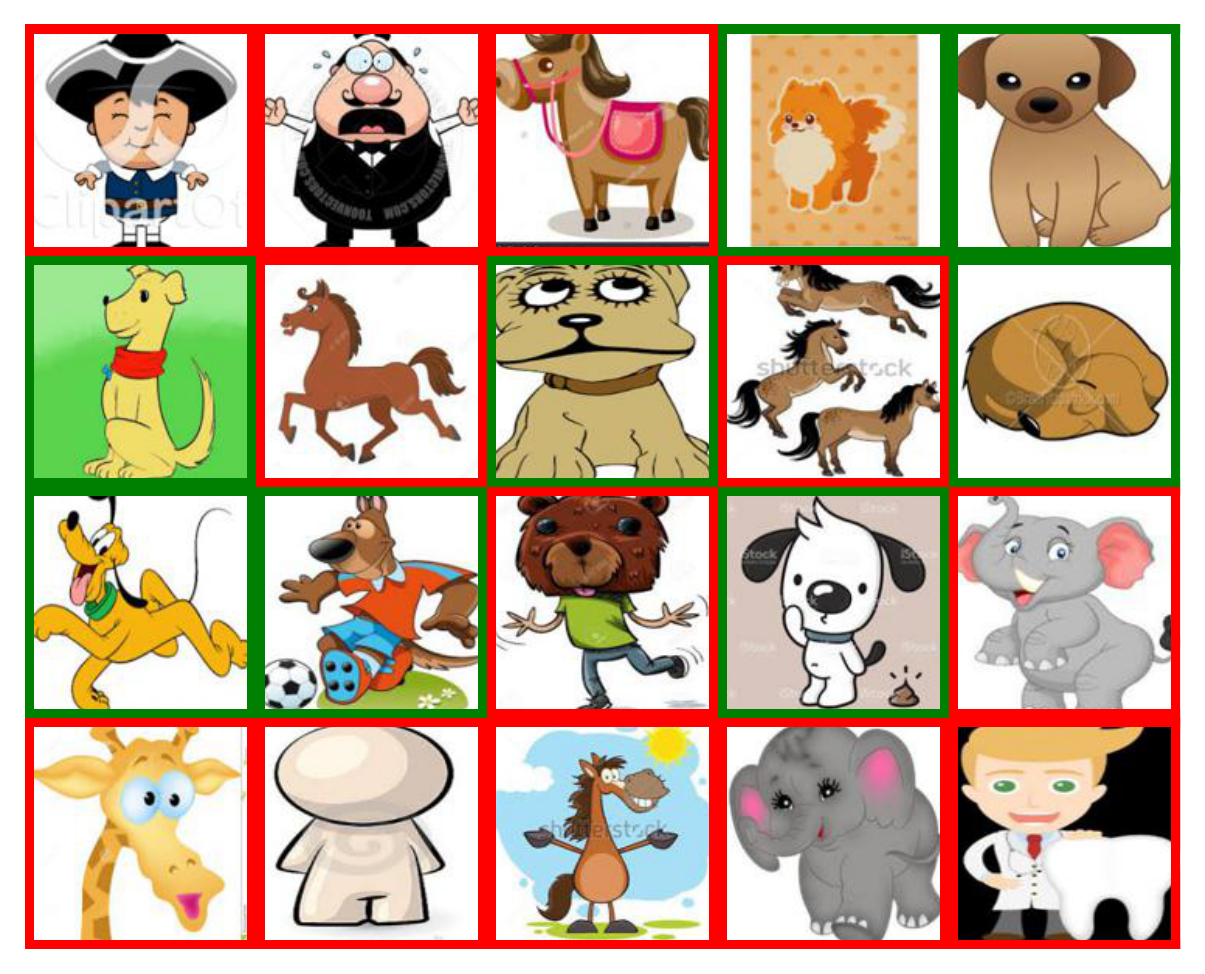}
  \end{minipage}
  \hfill
  \begin{minipage}[b]{0.24\textwidth}
    \includegraphics[width=\textwidth]{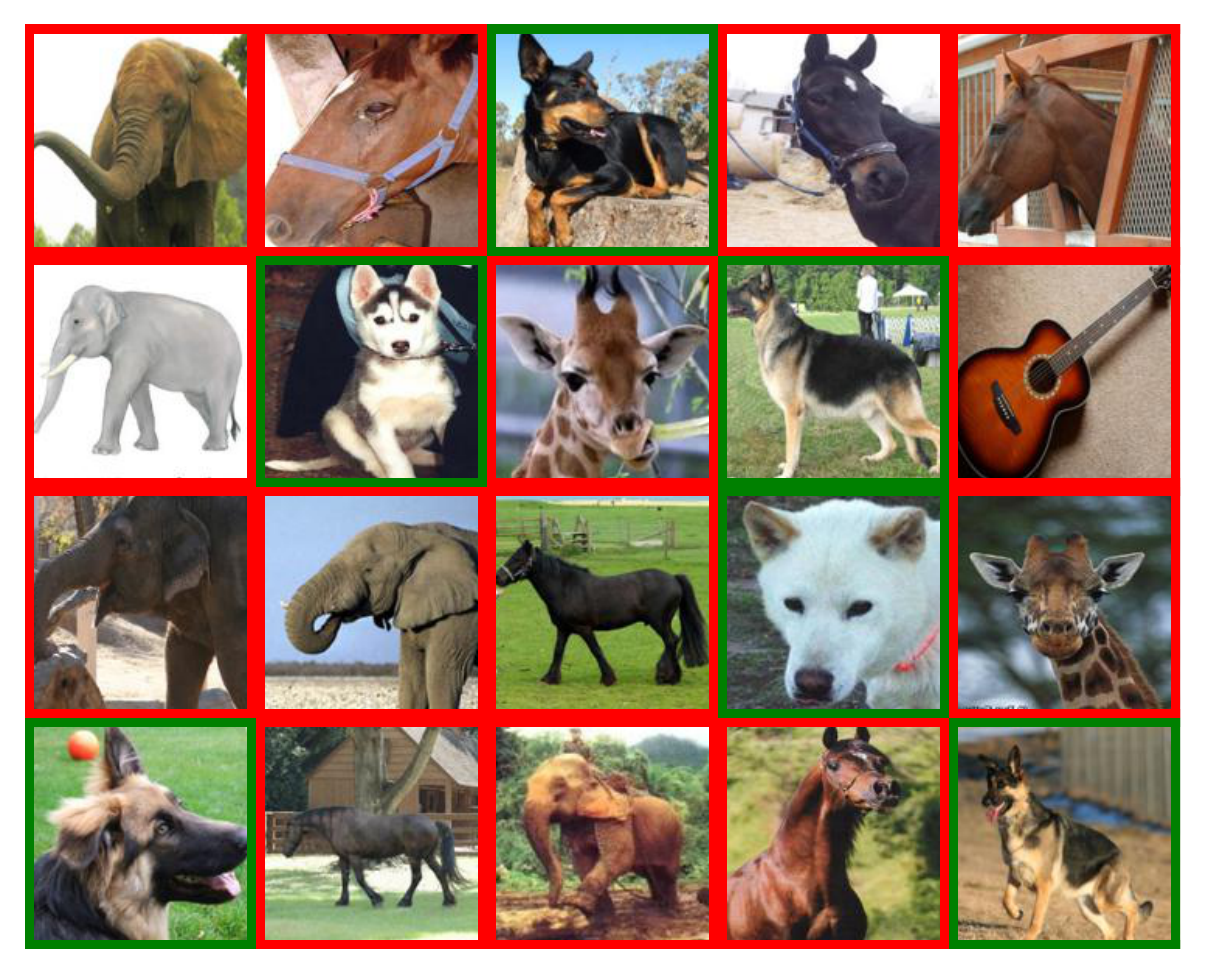}
  \end{minipage}
  \hfill
  \begin{minipage}[b]{0.24\textwidth}
    \includegraphics[width=\textwidth]{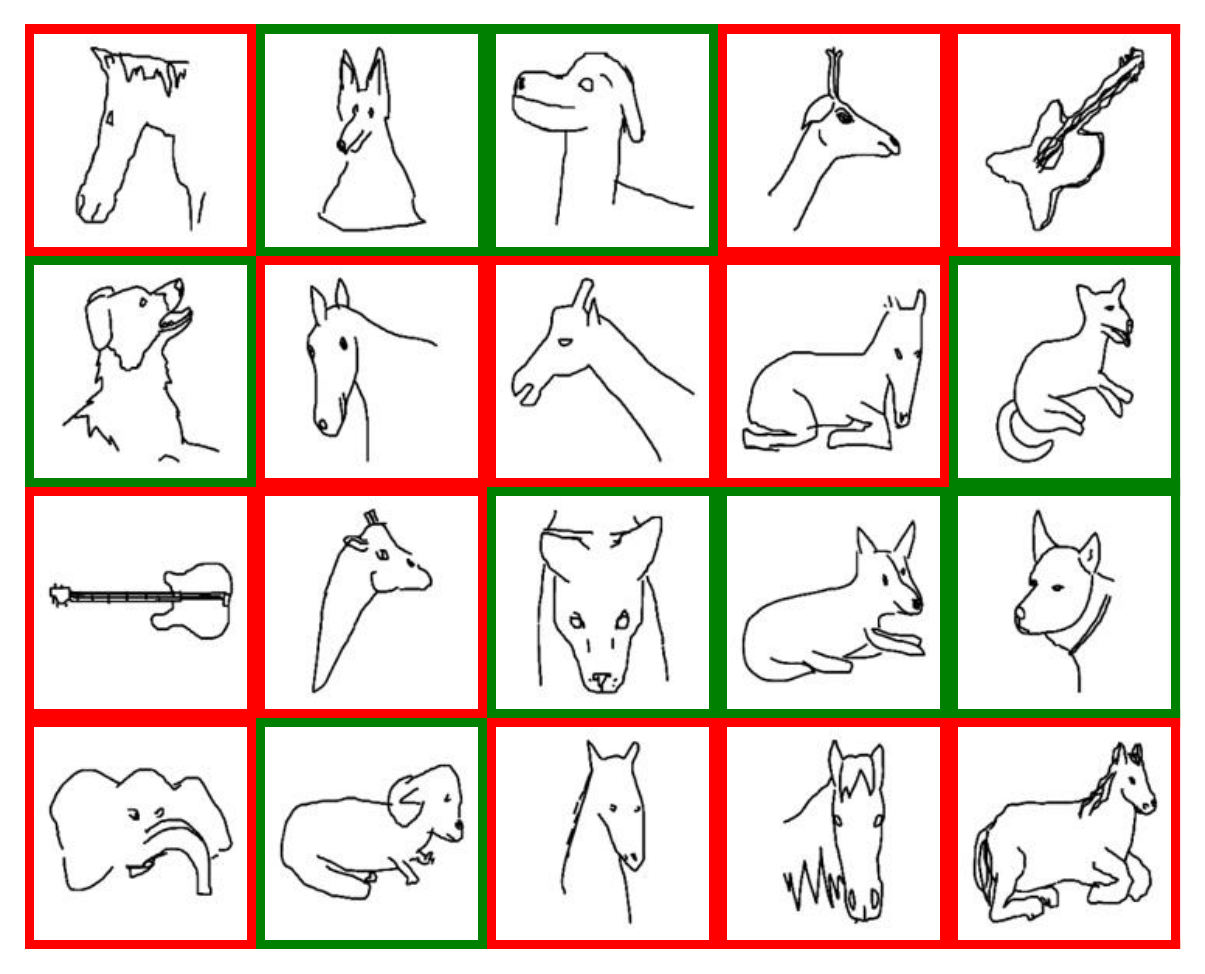}
  \end{minipage}
\hspace*{\fill}

\caption{Clusters corresponding to the PACS ``Dog'' class from Art, Cartoon, Photo and Sketch domains.}

\label{fig:tsne}
\end{figure}

\begin{figure}
\centering
\hspace*{\fill}
\begin{minipage}[b]{0.24\textwidth}
    \hspace{-0.1cm}
    \includegraphics[width=\textwidth]{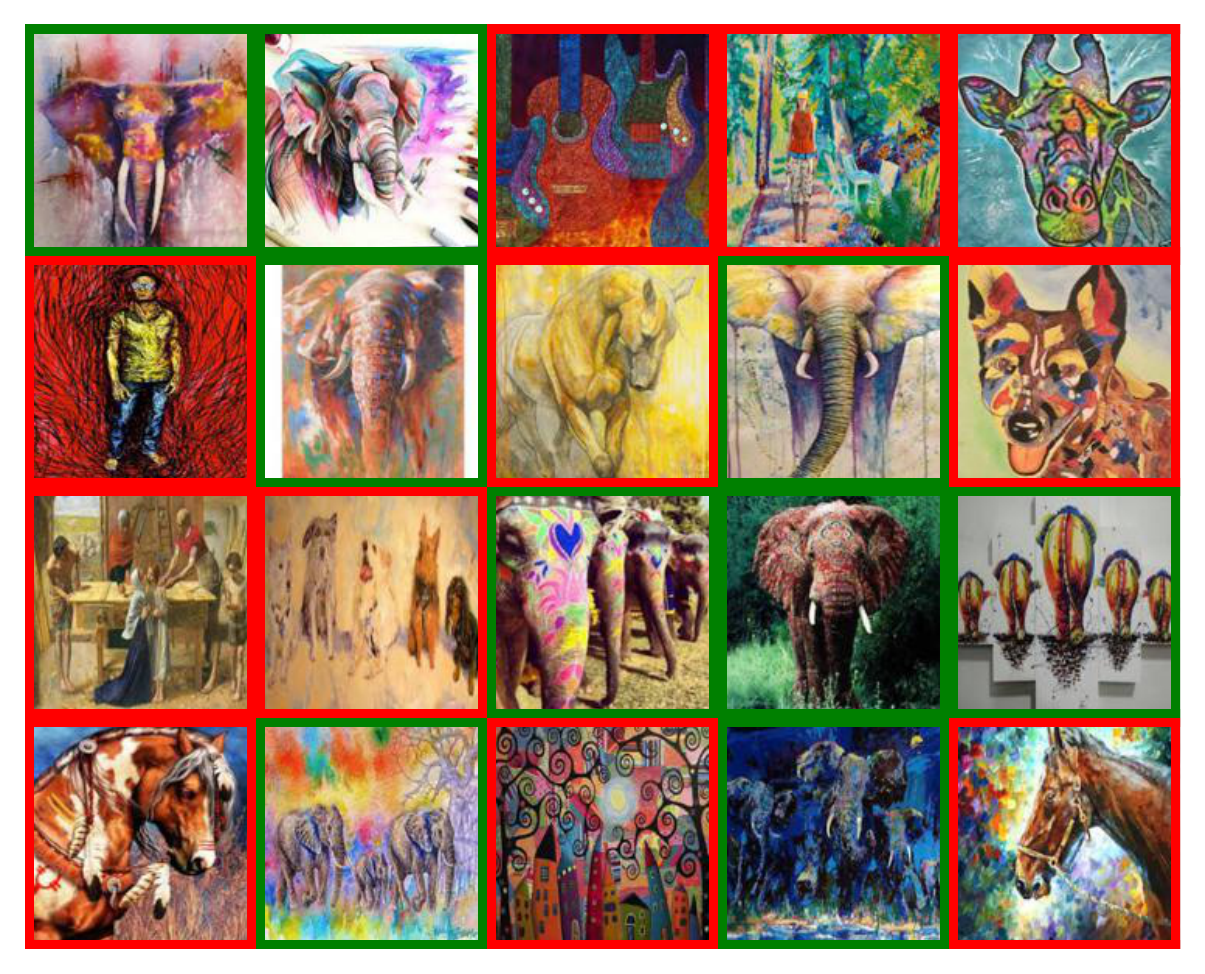}
  \end{minipage}
  \hfill
  \begin{minipage}[b]{0.24\textwidth}
    \includegraphics[width=\textwidth]{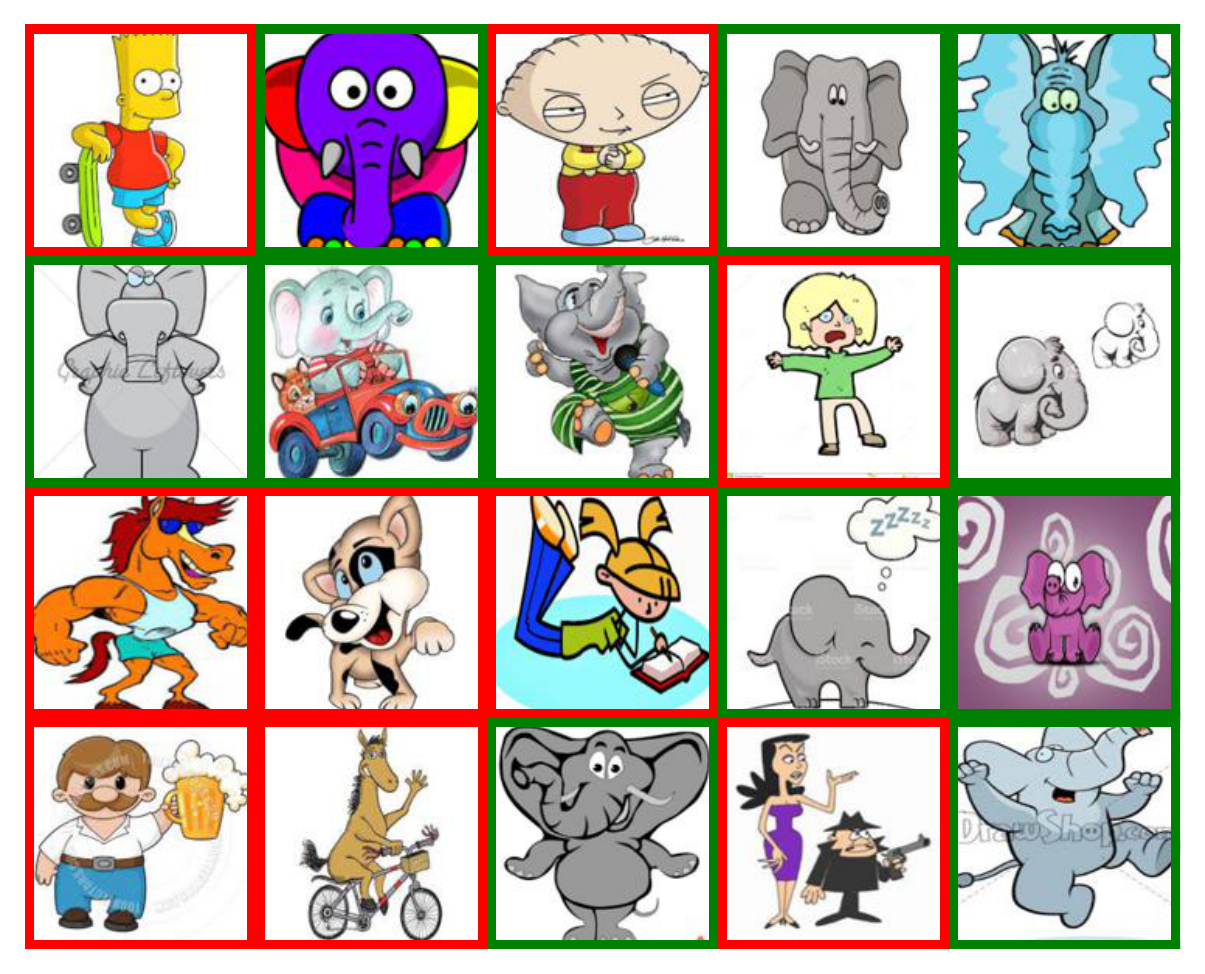}
  \end{minipage}
  \hfill
  \begin{minipage}[b]{0.24\textwidth}
    \includegraphics[width=\textwidth]{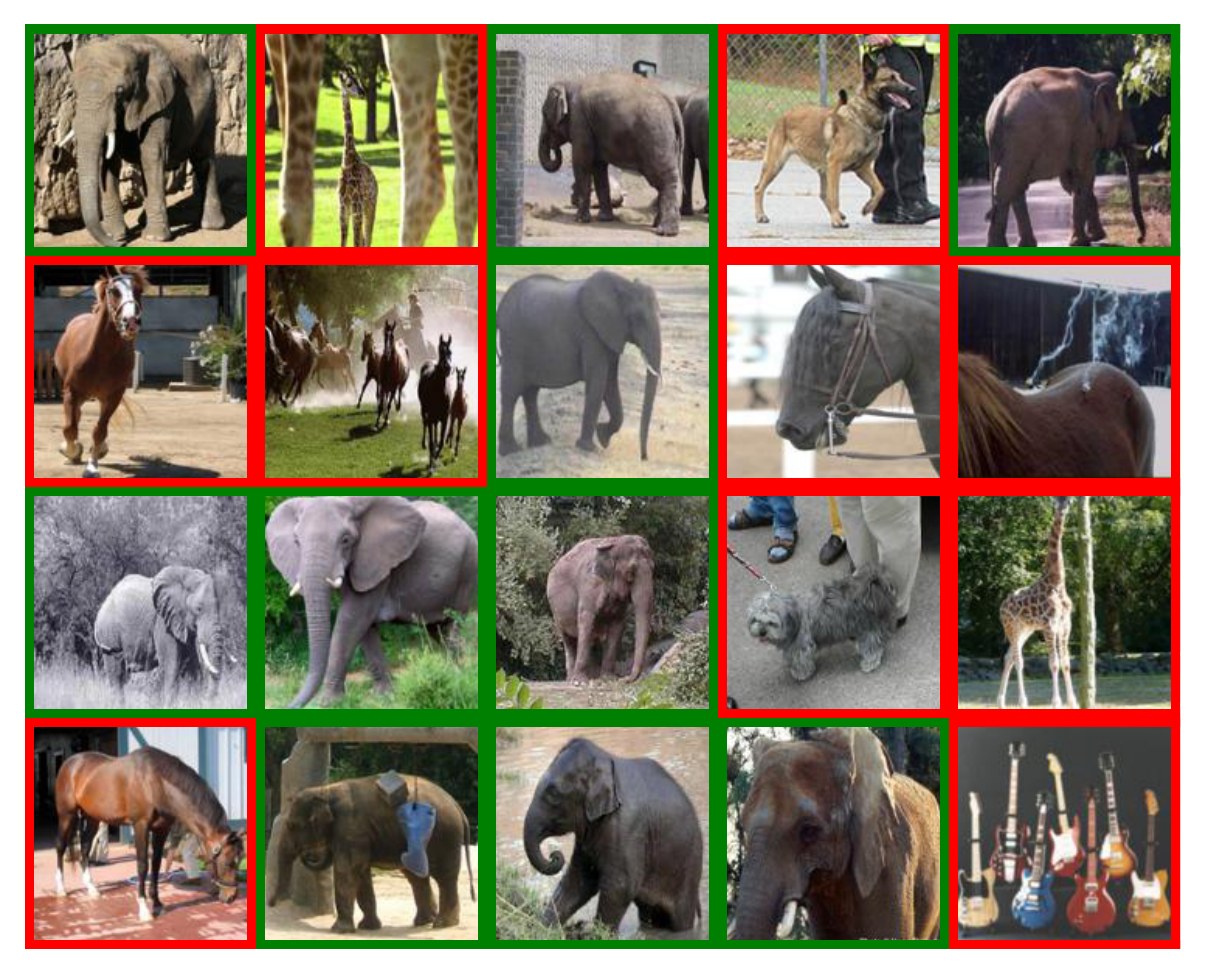}
  \end{minipage}
  \hfill
  \begin{minipage}[b]{0.24\textwidth}
    \includegraphics[width=\textwidth]{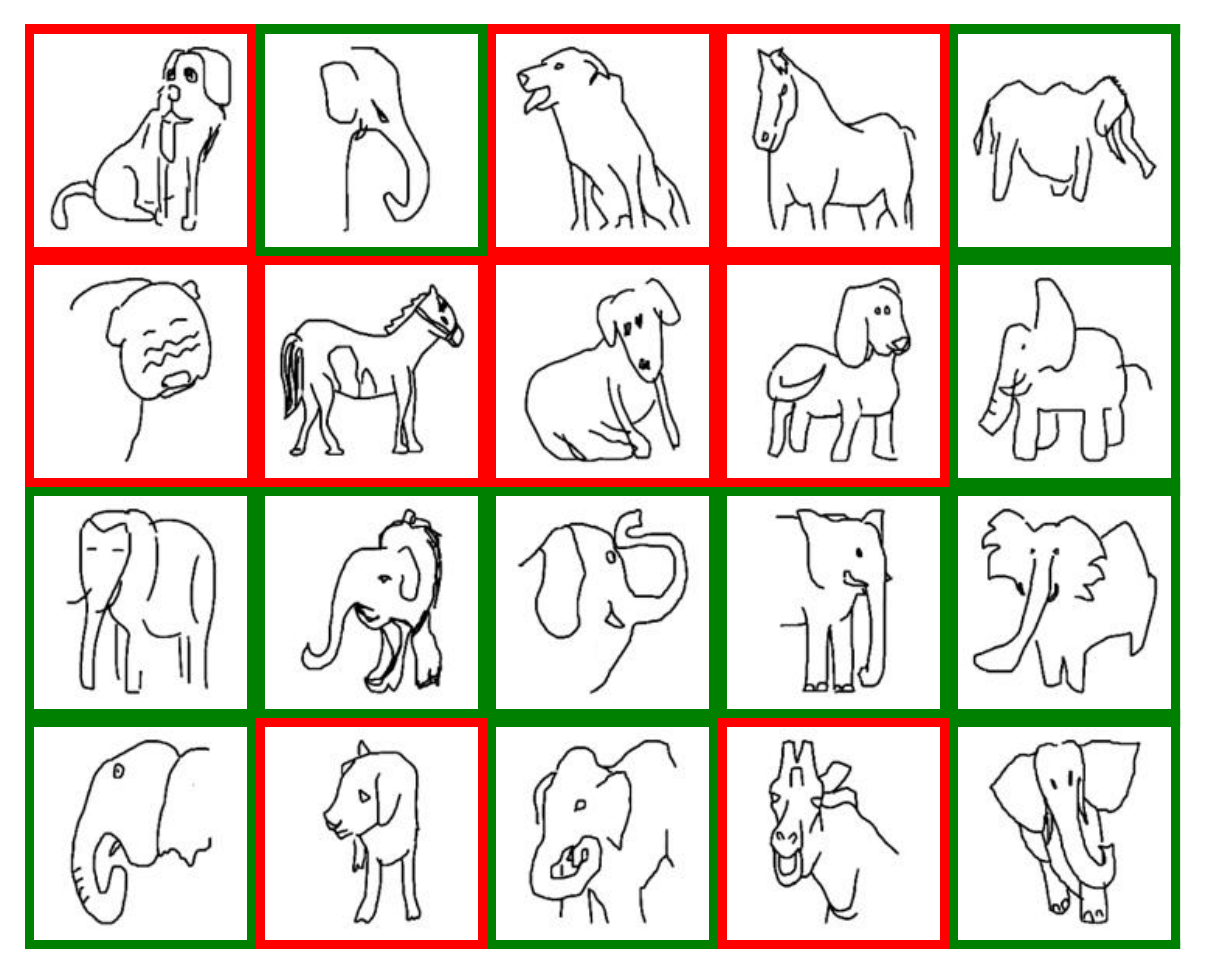}
  \end{minipage}
\hspace*{\fill}

\caption{Clusters corresponding to the PACS ``Elephant'' class from Art, Cartoon, Photo and Sketch domains.}

\label{fig:tsne}
\end{figure}

\begin{figure}
\centering
\hspace*{\fill}
\begin{minipage}[b]{0.24\textwidth}
    \hspace{-0.1cm}
    \includegraphics[width=\textwidth]{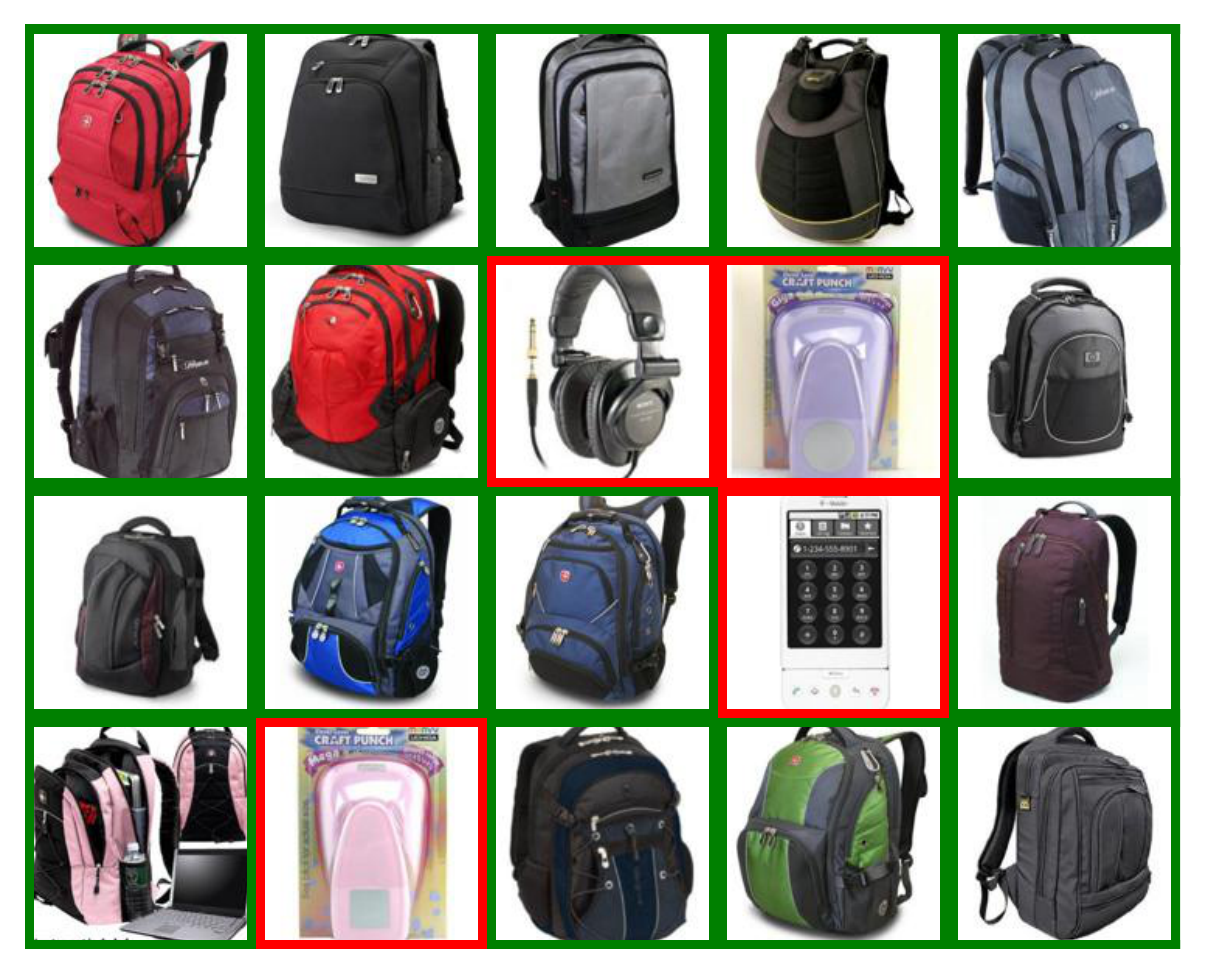}
  \end{minipage}
  \hfill
  \begin{minipage}[b]{0.24\textwidth}
    \includegraphics[width=\textwidth]{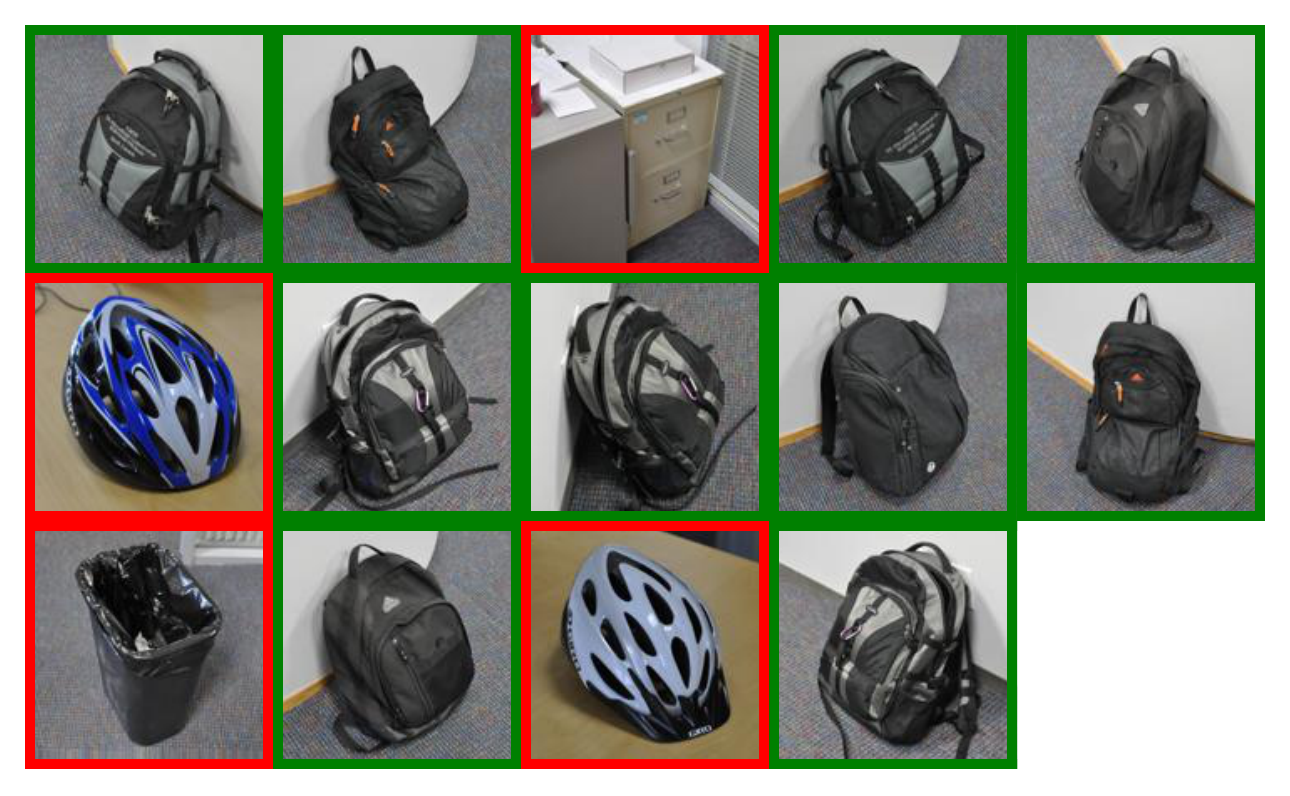}

  \end{minipage}
  \hfill
  \begin{minipage}[b]{0.24\textwidth}
    \includegraphics[width=\textwidth]{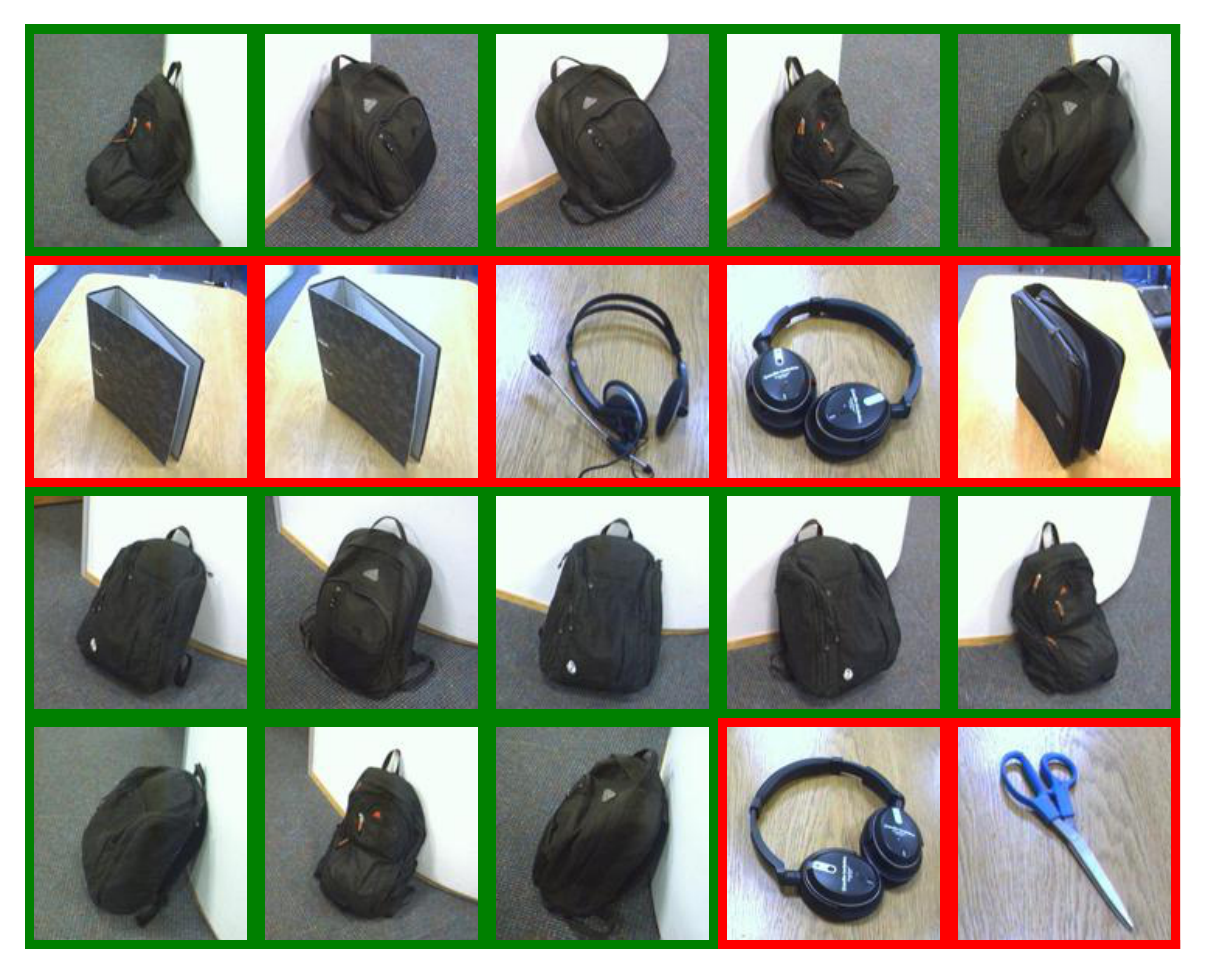}
  \end{minipage}
\hspace*{\fill}
\caption{Clusters corresponding to the Office31 ``Back Pack'' class from Amazon, DSLR and Webcam domains.}

\label{fig:tsne}
\end{figure}

\begin{figure}
\centering
\hspace*{\fill}
\begin{minipage}[b]{0.24\textwidth}
    \hspace{-0.1cm}
    \includegraphics[width=\textwidth]{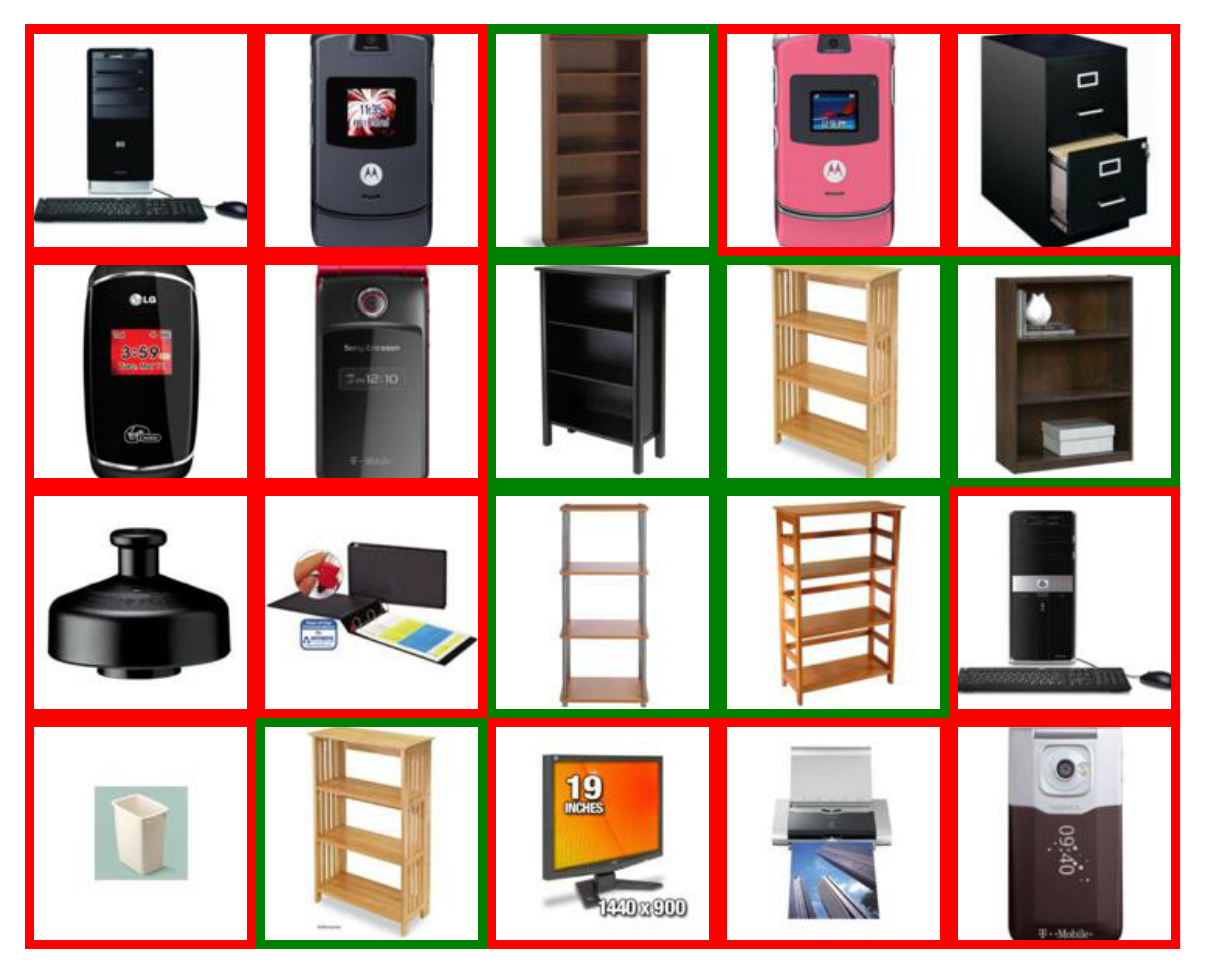}
  \end{minipage}
  \hfill
  \begin{minipage}[b]{0.24\textwidth}
    \includegraphics[width=\textwidth]{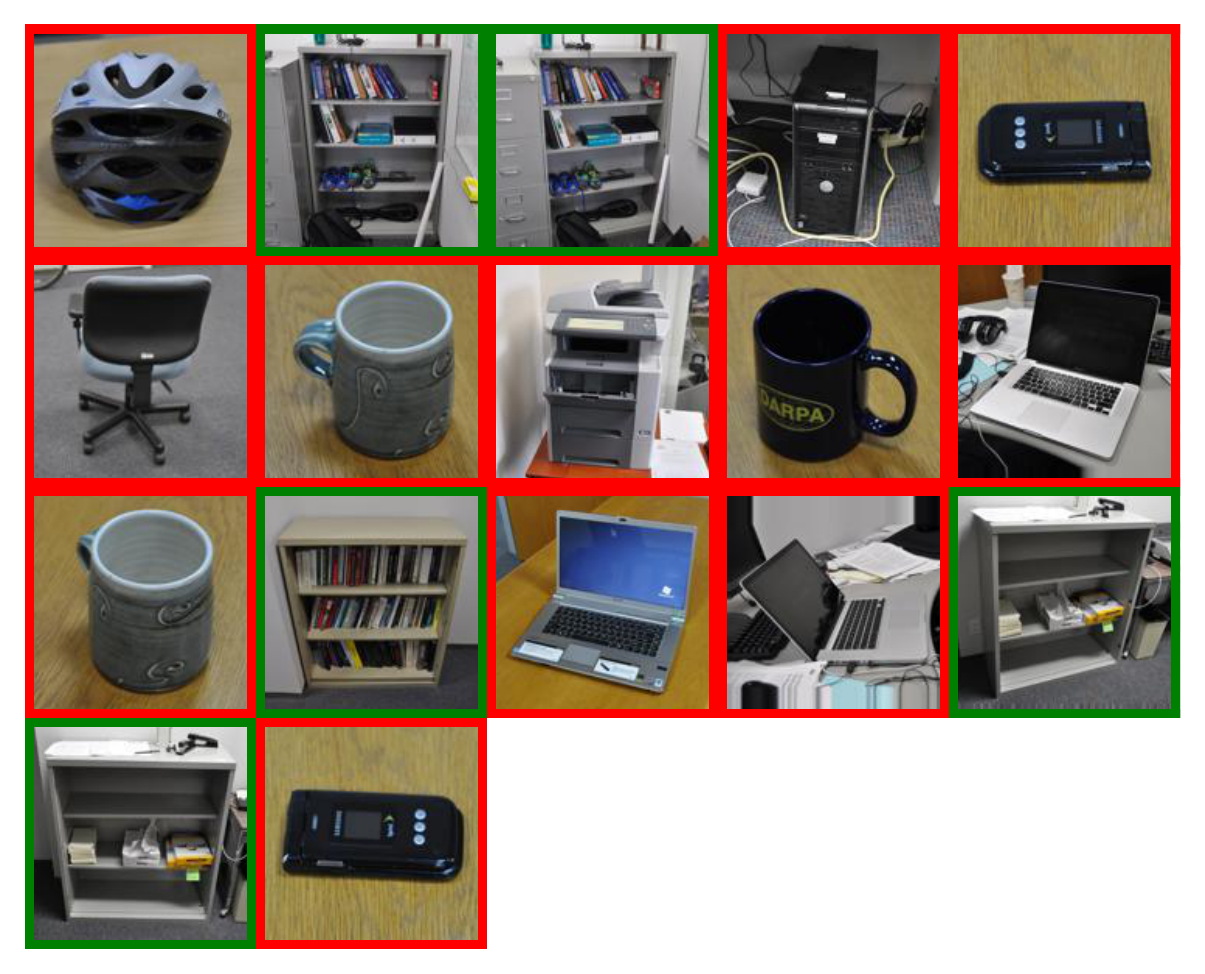}
  \end{minipage}
  \hfill
  \begin{minipage}[b]{0.24\textwidth}
    \includegraphics[width=\textwidth]{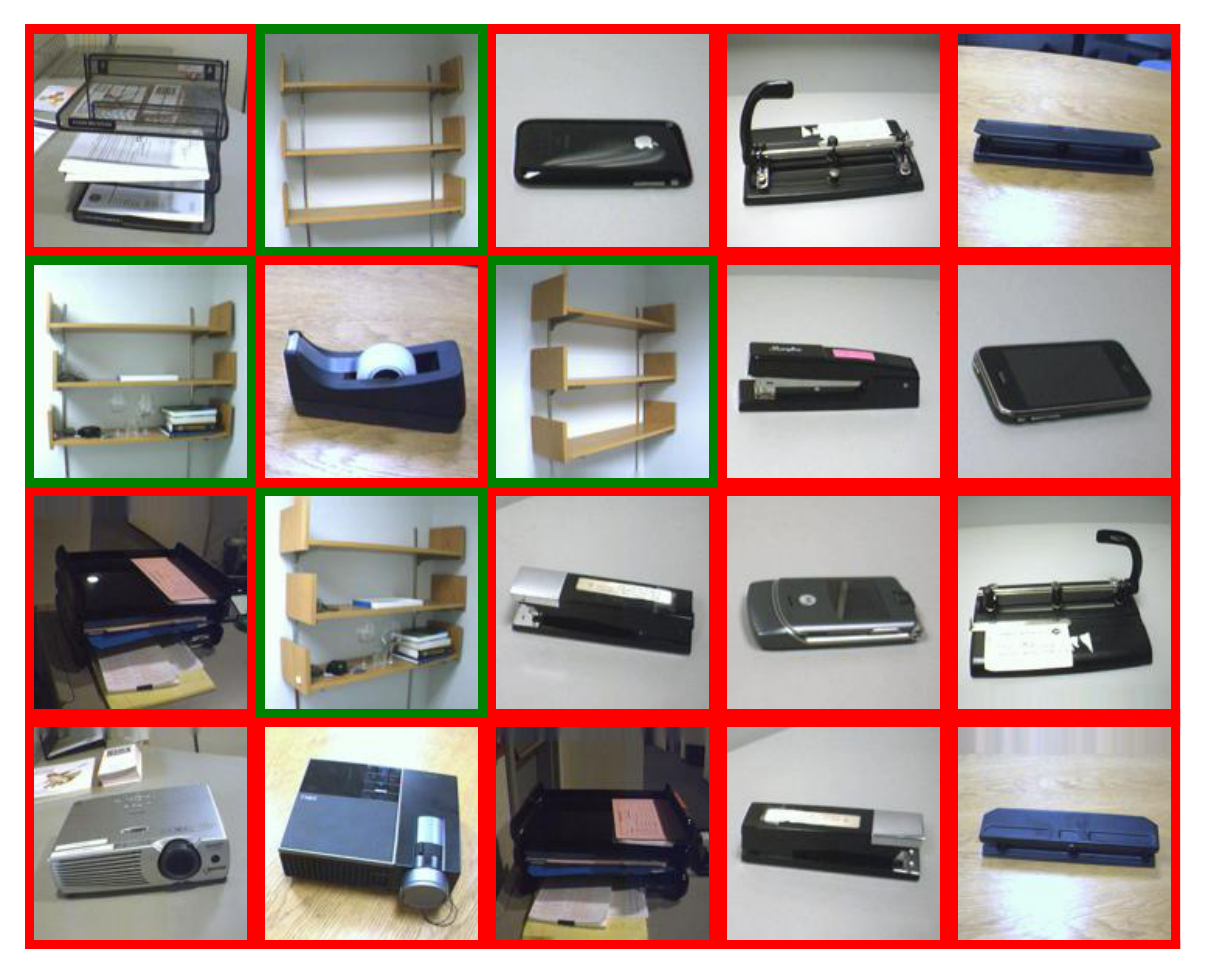}
  \end{minipage}
\hspace*{\fill}

\caption{Clusters corresponding to the Office31 ``Bookshelf'' class from Amazon, DSLR and Webcam domains.}

\label{fig:tsne}
\end{figure}

\begin{figure}
\centering
\hspace*{\fill}
\begin{minipage}[b]{0.24\textwidth}
    \hspace{-0.1cm}
    \includegraphics[width=\textwidth]{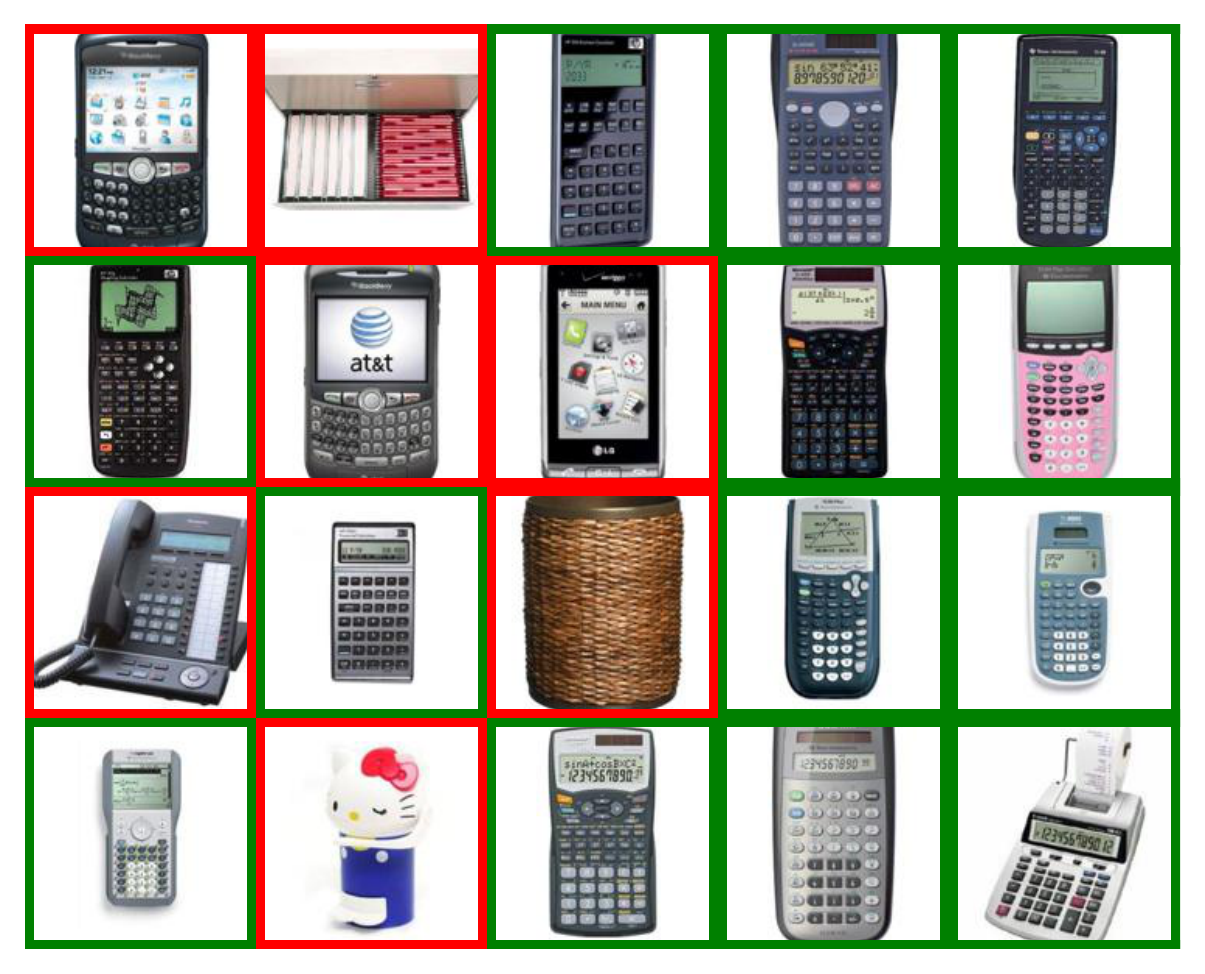}
  \end{minipage}
  \hfill
  \begin{minipage}[b]{0.24\textwidth}
    \includegraphics[width=\textwidth]{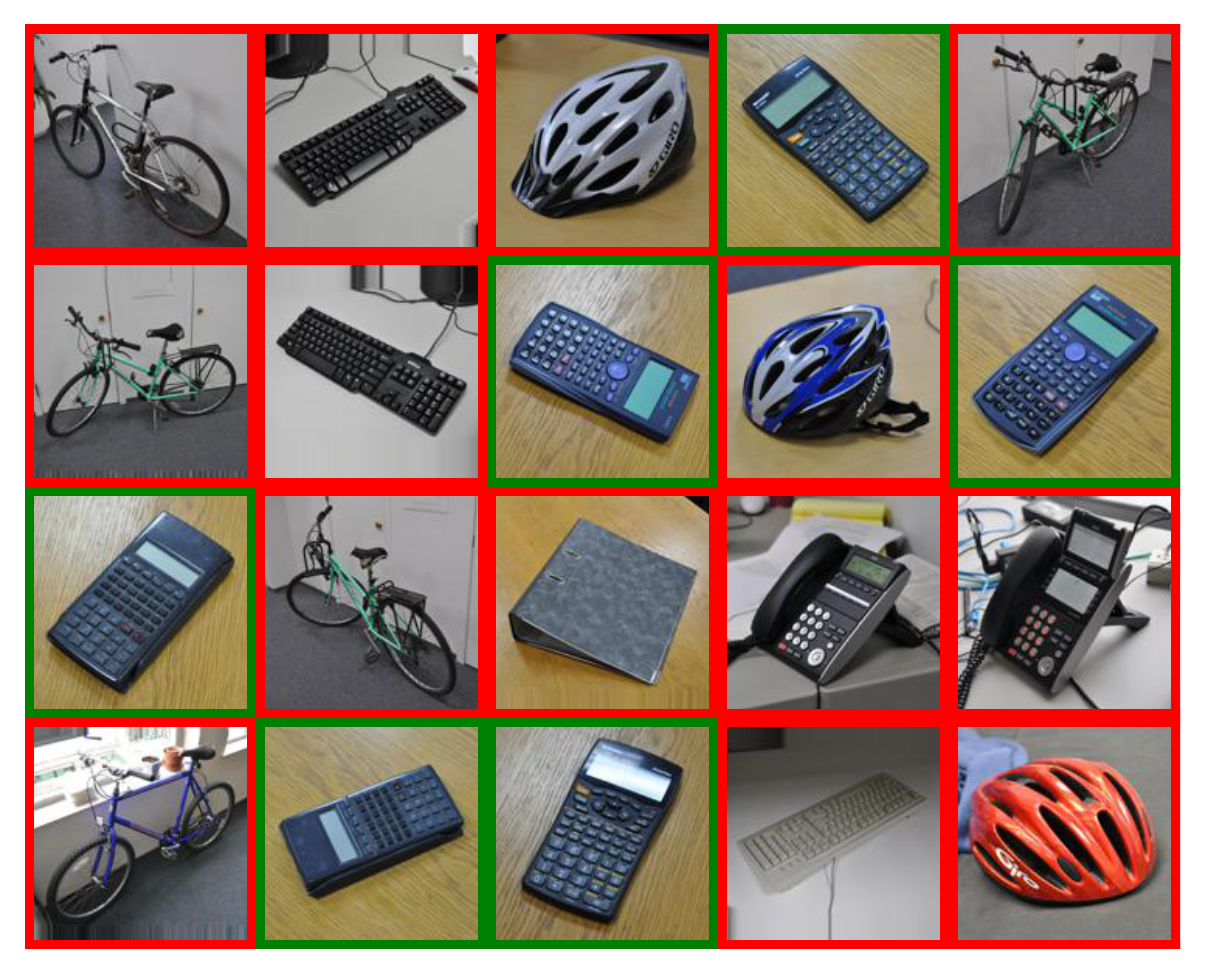}
  \end{minipage}
  \hfill
  \begin{minipage}[b]{0.24\textwidth}
    \includegraphics[width=\textwidth]{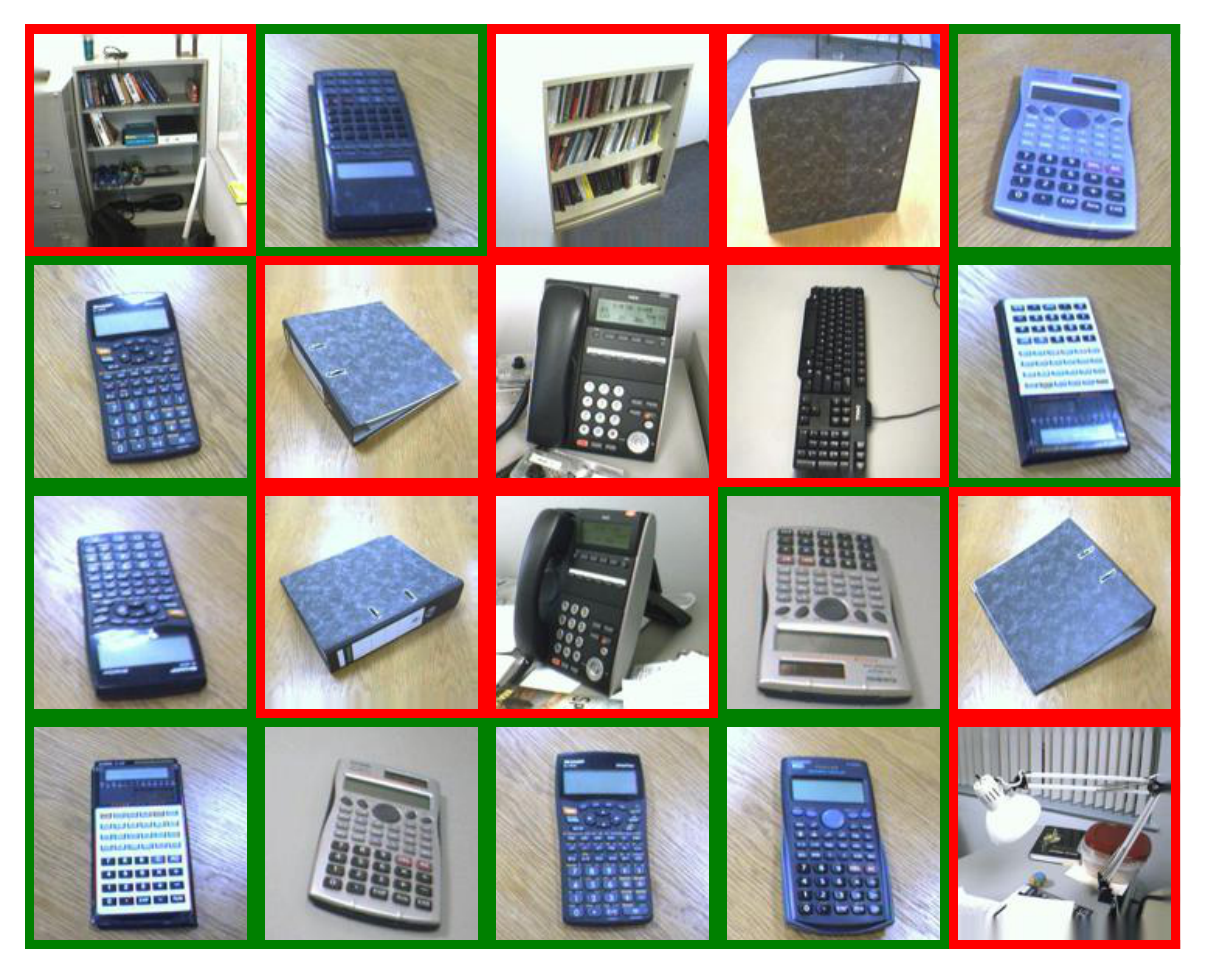}
  \end{minipage}
\hspace*{\fill}

\caption{Clusters corresponding to the Office31 ``Calculator'' class from Amazon, DSLR and Webcam domains.}

\label{fig:tsne}
\end{figure}

\begin{figure}
\centering
\hspace*{\fill}
\begin{minipage}[b]{0.24\textwidth}
    \hspace{-0.1cm}
    \includegraphics[width=\textwidth]{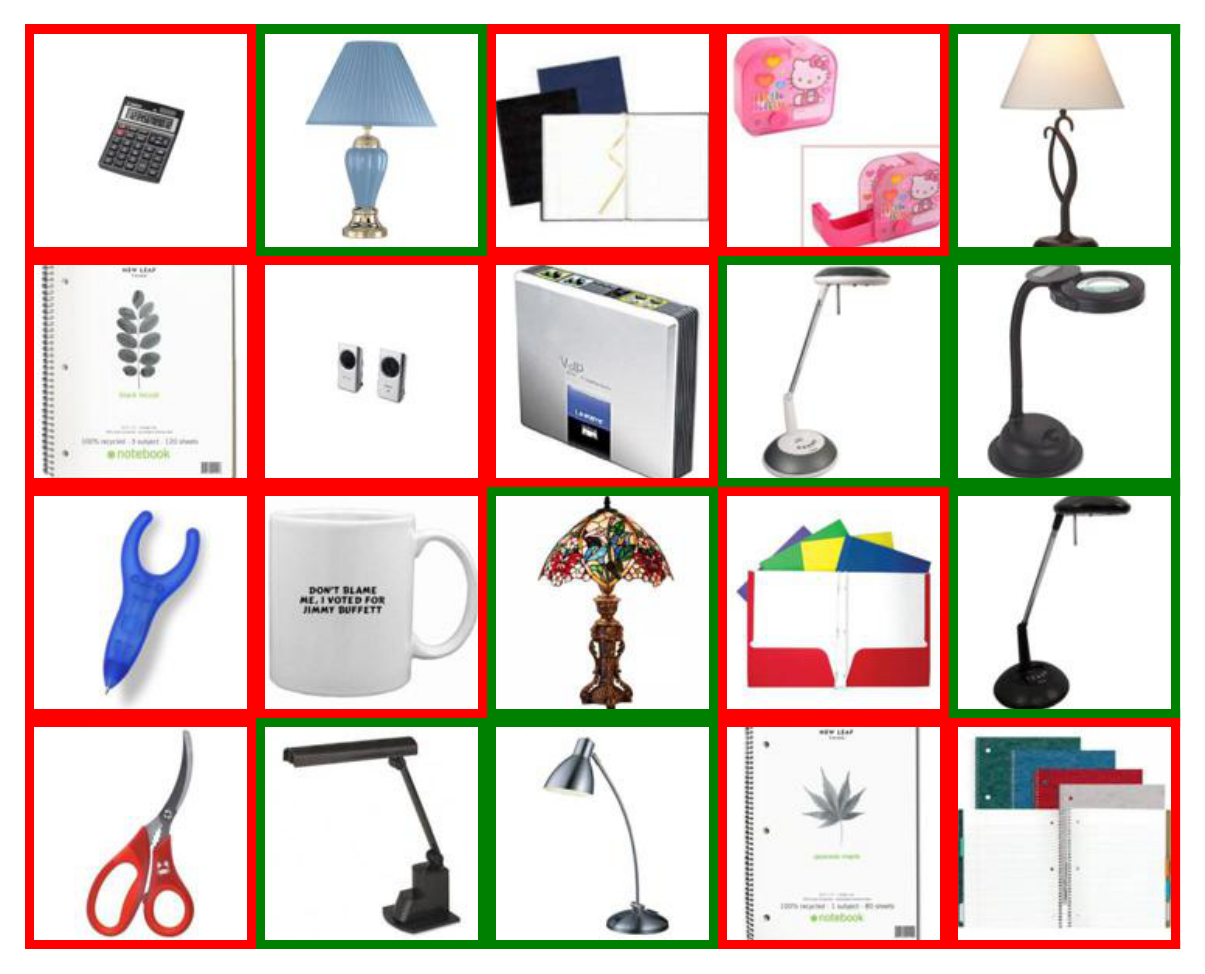}
  \end{minipage}
  \hfill
  \begin{minipage}[b]{0.24\textwidth}
    \includegraphics[width=\textwidth]{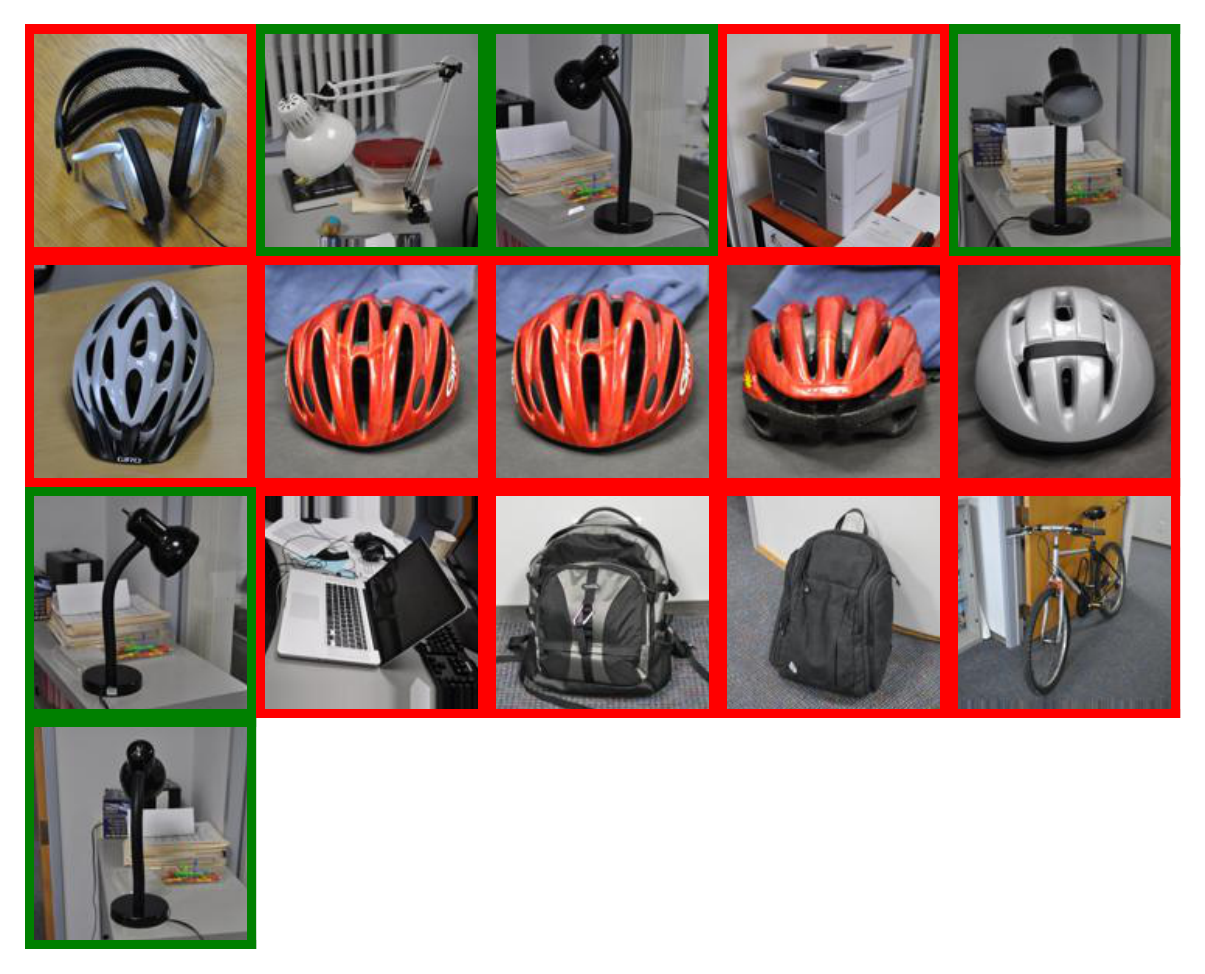}
  \end{minipage}
  \hfill
  \begin{minipage}[b]{0.24\textwidth}
    \includegraphics[width=\textwidth]{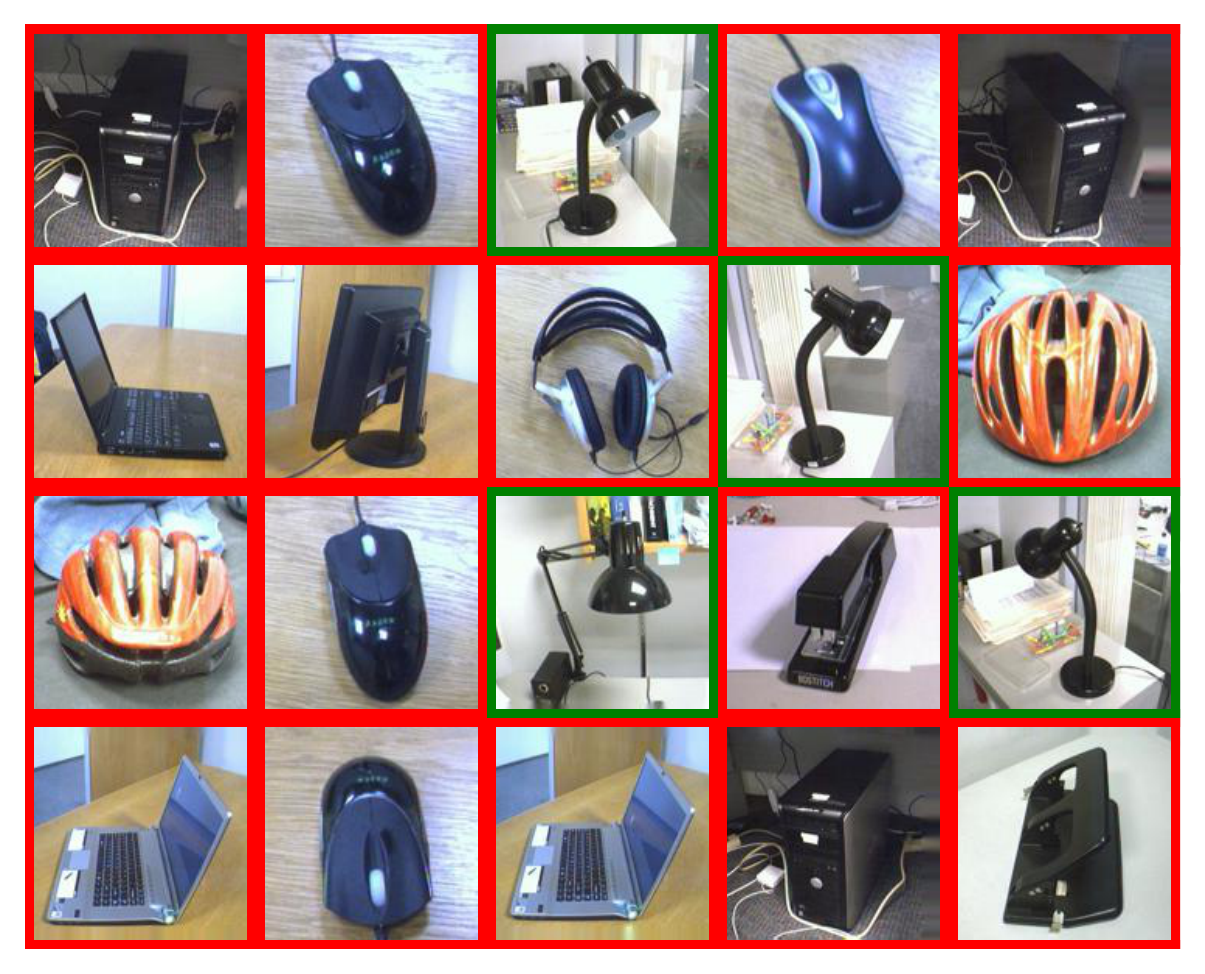}
  \end{minipage}
\hspace*{\fill}

\caption{Clusters corresponding to the Office31 ``Desk Lamp'' class from Amazon, DSLR and Webcam domains.}

\label{fig:tsne}
\end{figure}

\clearpage
%
%
\bibliographystyle{splncs04}
\bibliography{egbib}